\documentclass[journal,letterpaper,twocolumn]{IEEEtran}
\usepackage{amsmath,epsfig}
\usepackage{subfigure}
\usepackage{multirow}
\usepackage{url}
\title{Iterative Bilateral Filtering of Polarimetric SAR Data}
\author{Olivier D'Hondt, St\'{e}phane Guillaso and Olaf Hellwich
\thanks{The authors are with the Technische Universit\"{a}t Berlin (TUB), Computer Vision and Remote Sensing Group,
Franklinstrasse 28/29, D-10587 Berlin, Germany (e-mail : olivier.dhondt@tu-berlin.de, stephane.guillaso@tu-berlin.de, olaf.hellwich@tu-berlin.de)}%
\thanks{Color versions of figures are available in the online version of the manuscript.}%
}

\begin{document}
\maketitle
\begin{abstract}
  In this paper, we introduce an iterative speckle filtering method for polarimetric SAR (PolSAR) images based on the bilateral filter. To locally adapt to the spatial structure of images, this filter relies on pixel similarities in both spatial and radiometric domains. To deal with polarimetric data, we study the use of similarities based on a statistical distance called Kullback-Leibler divergence as well as two geodesic distances on Riemannian manifolds. To cope with speckle, we propose to progressively refine the result thanks to an iterative scheme. Experiments are run over synthetic and experimental data. First, simulations are generated to study the effects of filtering parameters in terms of polarimetric reconstruction error, edge preservation and smoothing of homogeneous areas. Comparison with other methods shows that our approach compares well to other state of the art methods in the extraction of polarimetric information and shows superior performance for edge restoration and noise smoothing. The filter is  then applied to experimental data sets from ESAR and FSAR sensors (DLR) at L-band and S-band, respectively. These last experiments show the ability of the filter to restore structures such as buildings and roads and to preserve boundaries between regions while achieving a high amount of smoothing in homogeneous areas.
\end{abstract}
\begin{IEEEkeywords}
Radar polarimetry, Image denoising, Covariance matrix, Statistical analysis, Parameter extraction.
\end{IEEEkeywords}
\section{Introduction}
\label{sec:intro}
\IEEEPARstart{S}{peckle} reduction of polarimetric synthetic aperture radar (PolSAR) data is often necessary to improve the outcome of applications such as physical parameter estimation, terrain classification and object detection, among others.
Polarimetric SAR systems provide information related to the backscattering mechanisms resulting of the interaction between a polarized electromagnetic wave and physical media. However, in the case of many natural media, due to the random distribution of the elementary scatterers present inside a resolution cell, phase and intensity cannot be predicted in a deterministic way \cite{Goodman1976}. The effect of the complex interactions between wave and scatterers is a pixel-to-pixel variability called speckle, that compromises the interpretation of the data. Though resulting of a physical mechanism, speckle has to be described as a signal dependent noise and to be handled in a statistical framework.

PolSAR systems provide images of complex vectors called target vectors. Due to the random properties of distributed targets, the coherent information carried by these vectors cannot be used directly and the data has to be processed in an incoherent way. Consequently, the covariance matrix of target vectors has to be estimated by computing an average over a set of independent samples. It is then possible to perform incoherent polarimetric decomposition methods \cite{Cloude1996} that allow a physical interpretation in terms of scattering mechanisms. Increasing the number of samples involved in the average allows to reduce speckle provided the statistical properties of the data are spatially stationary. Otherwise, pixels with different underlying scattering mechanisms are mixed, resulting in a loss of spatial resolution. Moreover, image statistics can be described by the fully-developed speckle model \cite{Goodman1976} only for distributed targets. If a strong return from a single scatterer dominates the overall response, the signal is better approximated by a deterministic target. In this case, averaging may be avoided and coherent decompositions can be directly applied (see for example \cite{Lee2009}, p. 214) to retrieve the polarimetric properties. 

The challenge faced by speckle reduction methods is then to perform an adaptation to the local structure of the image so as to reduce speckle variability without altering the polarimetric and spatial properties of the image. 
Thus, homogeneous areas have to be smoothed while boundaries between such regions have to be preserved. As for deterministic targets, they have to be left unfiltered.

Several adaptive methods have already been proposed for speckle reduction of PolSAR data. The popular refined Lee filter \cite{Lee1999} is based on a locally linear minimum mean square error assumption and performs matrix averaging so as to preserve the polarimetric properties of the signal. For a better preservation of spatial edges, a set of oriented windows is used. The refined Lee filter is computationally inexpensive and leads to good results as long as the spatial structures in the image are locally aligned with one of the oriented windows.

A region growing based method called intensity-driven adaptive-neighborhood (IDAN) allows an adaptation to the local structure of the image by extending Lee's minimum mean square approach to spatially adaptive neighborhoods \cite{Vasile2006}. However, only intensity information is considered in the selection of pixels in the neighborhood. 

A model based filter named MB-PolSAR \cite{Lopez-Martinez2008} allows the independent filtering of the matrix elements by considering an additive-multiplicative multidimensional speckle model. Although taking the full polarimetric information into account, this last method does not take advantage of the spatial information present in edges. 

A variational method based on anisotropic diffusion was also proposed \cite{Foucher2006}, achieving a high equivalent number of looks while preserving spatial information. Due to the use of partial differential equations, such a method has a high computational cost and is prone to numerical instabilities. 

A recent work exploits a multi-resolution representation of the image called binary partition tree (BPT) to group pixels with similar polarimetric properties \cite{Alonso-Gonzalez2012}. 
Though achieving a high number of looks and ensuring the preservation of strong edges, this method only allows the estimation of the mean covariance on a region basis, making it similar to a segmentation procedure.

Other approaches extended the non-local means filter to PolSAR data \cite{Deledalle2010, Chen2011}. Non-local means computes a weighted average of pixels according to the similarities between local neighborhoods called patches. Although the patches may in theory be located in any part of the image, search is generally restricted to a window to reduce the computational cost. One drawback of this method is that other similar neighborhoods have to be present in the search window.

In this work, we propose to take advantage of the full polarimetric information contained in covariance matrices while preserving the spatial resolution by adapting the bilateral filter \cite{Tomasi1998} to the processing of polarimetric covariance matrices. This filter has been widely used in  image processing due to its simplicity and its remarkable performance in terms of edge preservation. The idea of the filter is to compute a weighted mean of pixels contained in a local window according to their spatial and radiometric proximity. The weights are determined by two Gaussian kernels operating in radiometric and spatial domains. Furthermore, it has been shown that the iterative version of the bilateral filter is closely related to anisotropic diffusion \cite{Durand2002}.
To deal with PolSAR data the bilateral weights have to be adapted to the Hermitian positive definite nature of covariance matrices. One obvious choice would be to replace the traditional Euclidean distance between vectors by the Frobenius norm of matrix differences. However, this would lead to poor results due to the presence of speckle. It is then more suitable to consider distances that are adapted to the statistical nature of speckle. Statistical distances such as the Kullback-Leibler distance \cite{Kullback1951} may then be considered.  Moreover, recent studies \cite{Pennec2006, Barbaresco2009} have shown that a Riemannian affine-invariant metric was a natural choice due to the manifold structure of such matrices. Therefore, we propose to investigate the use of such distances to compute the radiometric proximity between two covariance matrices. We also consider a simpler distance called log-Euclidean \cite{Arsigny2006} that has been successfully applied in the field of diffusion tensor imaging (DTI).

Section \ref{sec:Data} introduces the basic concepts leading to the statistical speckle model for PolSAR data. 
Section \ref{sec:Bilat} describes the bilateral filter for gray-level and color images. In section \ref{sec:PolSAR} we introduce the bilateral filtering of PolSAR data based on distances between covariance matrices. The expression of a matrix based filter is given and several choices of distances are discussed. We also introduce an iterative scheme to progressively refine the estimate.
Section \ref{sec:Experiments} deals with the validation of the filter on synthetic and experimental data. A quantitative performance evaluation allows to understand the effects of parameters in terms of edge preservation, quality of polarimetric reconstruction and amount of smoothing. The new method is then compared to other existing methods using the criteria mentioned above. The method is then validated over two experimental datasets. 
Finally, the paper is concluded in section \ref{sec:Conclusion}. 

\section{Polarimetric SAR data}
\label{sec:Data}
Polarimetric SAR systems measure the relation between the transmitted and received electromagnetic wave in two orthogonal polarizations in the form of a scattering matrix \cite{Lee2009}
\begin{equation}
\mathbf{S} =
\left[
\begin{array}{cc}
  S_{hh} & S_{hv}  \\
  S_{vh} & S_{vv}     
\end{array}
\right],
\label{eqn:scatmatrix}
\end{equation}
where $h$ and $v$ denote horizontal and vertical polarization, respectively. The reciprocity assumption in the mono-static case leads to $S_{hv} = S_{vh}$. For further analysis, it is convenient to represent the scattering information in the form of a target vector containing the elements of $\mathbf{S}$
\begin{equation}
	\mathbf{k}_l = [S_{hh}, \sqrt{2} S_{hv}, S_{vv}]^T,
\end{equation}
\begin{equation}
	\mathbf{k}_p = \frac{1}{\sqrt{2}}[S_{hh} + S_{vv}, S_{hh} - S_{vv}, 2S_{hv}]^T,
\end{equation}
where $l$ and $p$ denote the lexicographic and Pauli bases.

On distributed targets, due to the complex interactions between the wave and multiple scatterers inside the resolution cell, this information is generally considered in a statistical framework. Under the fully developed speckle hypothesis, the target vectors $\mathbf{k}$ follow a complex circular $d$-variate Normal distribution \cite{Goodman1963}
\begin{equation}
 p(\mathbf{k}) = \frac{1}{\pi^{d}|\mathbf{\Sigma}|} \exp{\left(-\mathbf{k}^{\dagger}\mathbf{\Sigma}^{-1}\mathbf{k}\right)},
\end{equation}
where $|.|$ is the determinant of a matrix and $\dagger$ represents the conjugate transpose of a complex vector.
This distribution is fully described by its covariance matrix $\mathbf{\Sigma}=E[\mathbf{k}\mathbf{k}^\dagger]$ that contains information about power and relative phase between polarimetric channels. If $\mathbf{k} = \mathbf{k}_l$ then $ \mathbf{\Sigma}=\mathbf{C}$ is called polarimetric covariance, and if $\mathbf{k} = \mathbf{k}_p$ then $\mathbf{\Sigma}=\mathbf{T}$ is called polarimetric coherency.

Unfortunately, $\mathbf\Sigma$ has to be estimated from the data over several independent samples, resulting in a loss of spatial resolution. The sample covariance is expressed as
\begin{equation}
  \mathbf{\widehat{\Sigma}} = \frac{1}{L}\sum_{i=1}^{L} \mathbf{k}_i\mathbf{k}_i^\dagger,
  \label{eqn:multilooking}
\end{equation}
and is obtained by an operation called \textit{multi-looking} that consists in taking the average of contiguous pixels. 

Under this Gaussian hypothesis, the multi-look covariance matrix follows a complex Wishart density \cite{Goodman1963}. The estimation accuracy can be improved by increasing the number of samples. Unfortunately, increasing this number is only possible on homogeneous areas where pixels arise from identical scattering mechanisms.

\section{The bilateral filter for gray-level and color images}
\label{sec:Bilat}
Let us consider a 2-D gray level image $I:\Omega \subset R^2 \rightarrow R$ where a pixel is referenced by its 2-D location $\mathbf{x}$. The original expression of the bilateral filter at $\mathbf{x_0}$ is \cite{Tomasi1998}
\begin{equation}
  F(\mathbf{x}_0) = \frac{1}{K} \int_{ \mathbf{x} \in\Omega} I(\mathbf{x})  f_s(\|\mathbf{x}-\mathbf{x}_0\|)  f_r\left(|I(\mathbf{x}) - I(\mathbf{x}_0)|\right)d \mathbf{x}
\label{eqn:bil_orig}
\end{equation}
with
\begin{equation}
	K = \int_{ \mathbf{x} \in\Omega} f_s(\|\mathbf{x}-\mathbf{x}_0\|)f_r\left(|I(\mathbf{x}) - I(\mathbf{x}_0))|\right)d \mathbf{x},
        \label{eqn:normalize}
\end{equation}
where the operator $\|.\|$ is the $L_2$ norm. Weighting functions $f_s$ and $f_r$ have to be chosen such as they attain a maximum in zero and tend to zero if their argument goes to infinity. In this work we adopt the functions defined in the original version of the filter \cite{Tomasi1998} where $f_s(u) = g_{\gamma_s}(u)$ and $f_r(u) = g_{\gamma_r}(u)$ with
\begin{equation}
	g_{\gamma}(u) = \exp(-\frac{u^2}{\gamma^2}).
\end{equation}
The parameter $\gamma_s$ controls the spatial extent of the filter and is comparable to the window size used in other techniques like the Lee filter \cite{Lee1999}. The parameter $\gamma_s$ controls the amount of filtering according to the radiometric proximity between two pixels.
The filter thus performs a local average of intensities, assigning weights to pixels depending on their spatial and radiometric similarity to the pixel located at $\mathbf{x_0}$. 

The filter is easily generalized to vectorial (\textit{e.g.} color) images $\mathbf{I} :\Omega \subset R^2 \rightarrow R^N
$ by replacing the absolute value by the $L_2$ norm in equation \eqref{eqn:bil_orig} and \eqref{eqn:normalize}. In practice, the filter is applied to discrete images and approximated by a weighted sum of the pixel values in a finite local window  $\mathcal{W}$ centered around $\mathbf{x_0}$
\begin{equation}
  \mathbf{F}(\mathbf{x_0}) = \sum_{ \mathbf{x_i} \in \mathcal{W}} w_i \mathbf{I}(\mathbf{x_i}),
  \label{eqn:bilat_discr}
\end{equation}
where the weights are defined by
\begin{equation}
  w_i = \frac{  f_s(\|\mathbf{x_i}-\mathbf{x_0}\|)  f_r\left(\| \mathbf{I}(\mathbf{x_i}) - \mathbf{I}(\mathbf{x_0})\|\right)}
  {\sum_{ \mathbf{x_i} \in \mathcal{W}}  f_s(\|\mathbf{x_i}-\mathbf{x_0}\|)  f_r\left(\| \mathbf{I}(\mathbf{x_i}) - \mathbf{I}(\mathbf{x_0})\|\right)}.
  \label{eqn:bilat_weights}
\end{equation}

\begin{figure}[t]
  \centering
  \subfigure[Structure 1]{
    \includegraphics[scale=0.17]{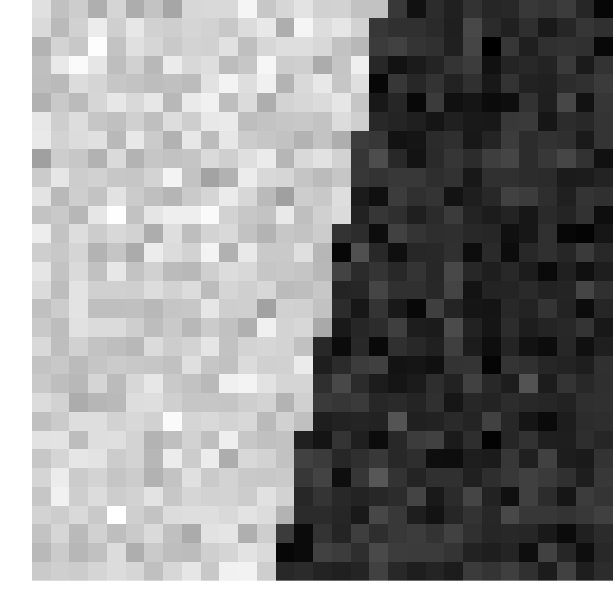}
    \label{sfg:struct1}
  }
  \subfigure[$f_s$]{
    \includegraphics[scale=0.17]{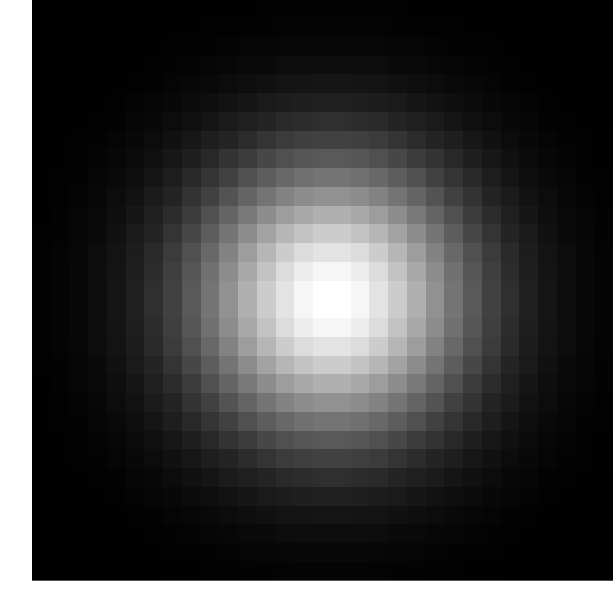}
    \label{sfg:fs1}
  }
  \subfigure[$f_r$]{
    \includegraphics[scale=0.17]{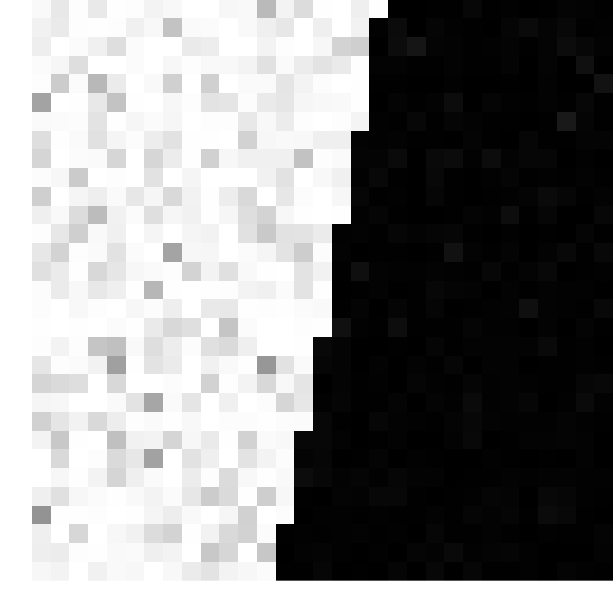}
    \label{sfg:fr1}
  }
  \subfigure[$f_s\times f_r$]{
    \includegraphics[scale=0.17]{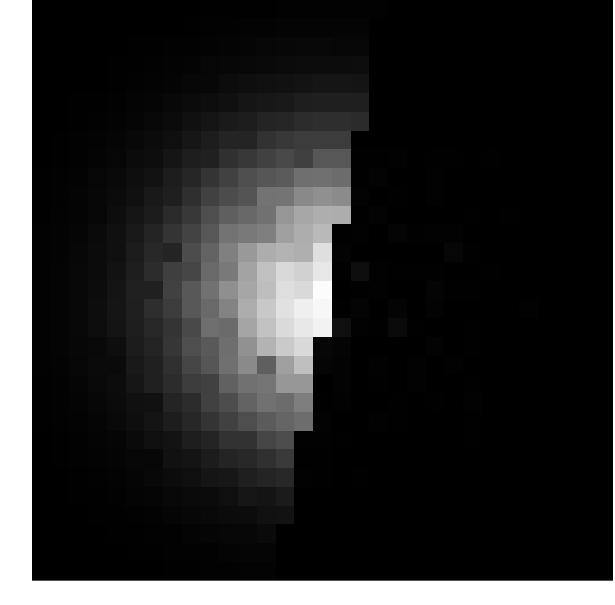}
    \label{sfg:w1}
  }

  \subfigure[Structure 2]{
    \includegraphics[scale=0.17]{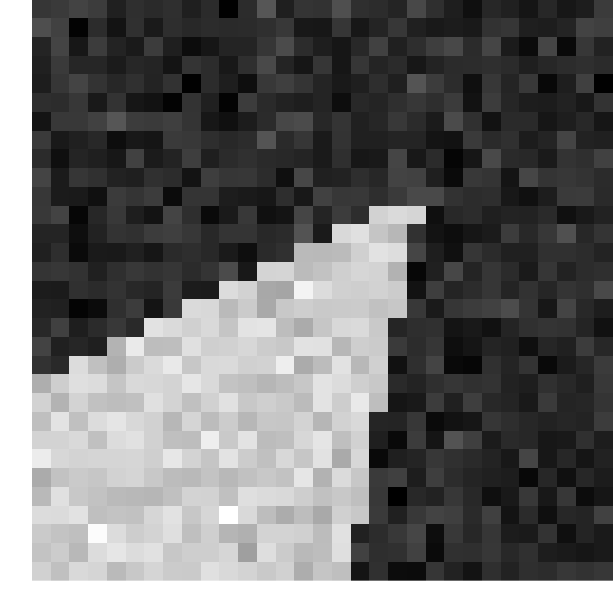}
    \label{sfg:struct2}
  }
  \subfigure[$f_s$]{
    \includegraphics[scale=0.17]{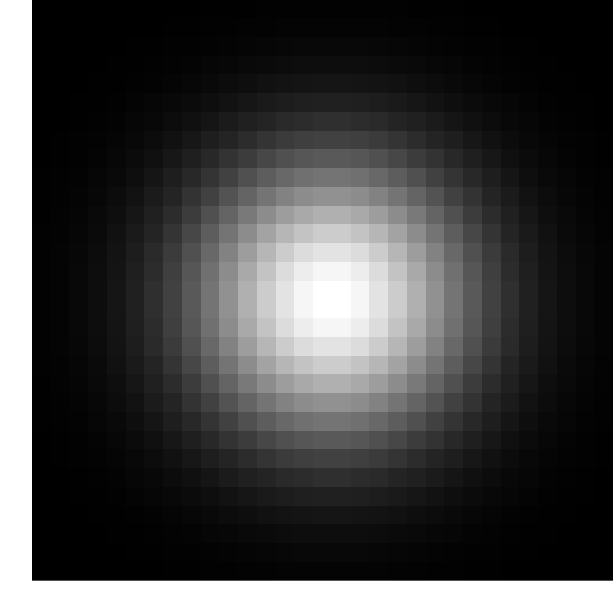}
    \label{sfg:fs2}
  }
  \subfigure[$f_r$]{
    \includegraphics[scale=0.17]{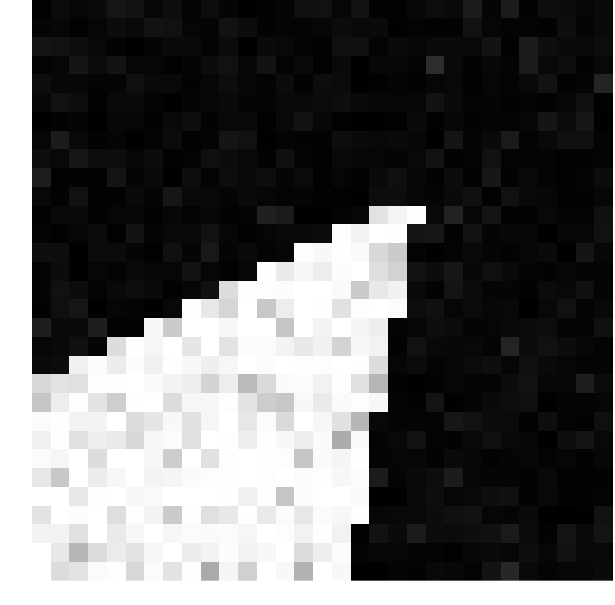}
    \label{sfg:fr2}
  }
  \subfigure[$f_s\times f_r$]{
    \includegraphics[scale=0.17]{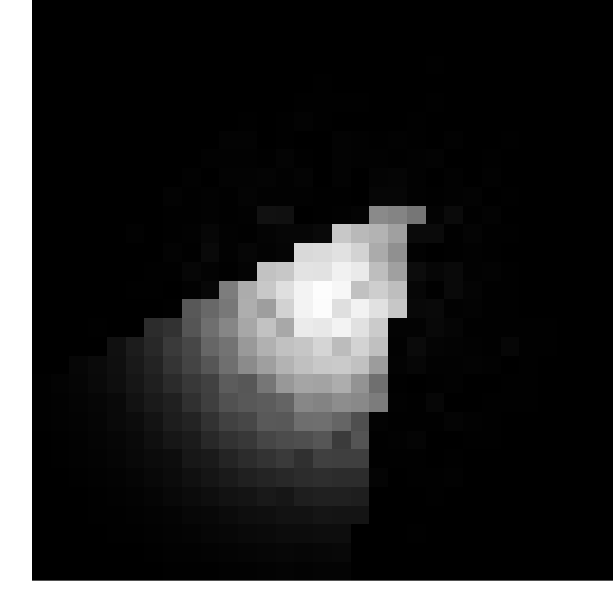}
    \label{sfg:w2}
  }

  \subfigure{
    \includegraphics[scale=0.35]{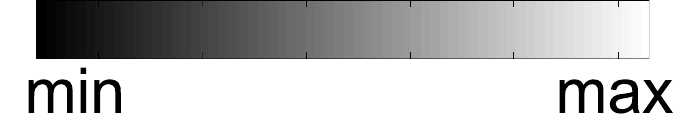}
  }
  \caption{Illustration showing the spatial adaptation capabilities of the bilateral filter for 2 examples of discontinuities in noisy images \subref{sfg:struct1}, \subref{sfg:struct2} assuming the pixel to filter belongs to the brightest area. The spatial weights $f_s$ \subref{sfg:fs1}, \subref{sfg:fs2} only depend on the spatial distance to the center of the window while the radiometric weights $f_r$ \subref{sfg:fr1}, \subref{sfg:fr2} adapt to the intensity difference with respect to the brightness of the central pixel. The bilateral weights are given by the product $f_s\times f_r$ \subref{sfg:w1}, \subref{sfg:w2}}
  \label{fig:bilatprinciple}
\end{figure}

Fig. \ref{fig:bilatprinciple} shows a visual interpretation of the filter behaviour for two examples of spatial structures in grey level images. As it may be observed, the strength of the bilateral filter is that it can adapt not only to straight edges but to any shape of discontinuity provided the edge is characterized by an abrupt intensity transition. For PolSAR data, this should constitute an advantage over other traditional filtering methods that use a discrete set of oriented windows. A drawback of this method is that it relies on the value of the central pixel that is uncertain because of noise. However, this limitation may be overcome by considering an iterative version of the filter as we show in the following. For an extensive survey on applications and variants of the bilateral filter, that is beyond the scope of this paper, the reader may refer to \cite{Paris2009}.

\section{Extension of the filter to PolSAR data}
\label{sec:PolSAR}


\subsection{Bilateral filtering of covariance matrices}
\label{sub:BilateralMatrix}

To be able to use the bilateral filter on PolSAR data, it has to handle polarimetric covariance matrices instead of vectors. Equation \eqref{eqn:bilat_discr} thus becomes a weighted sum of covariances
\begin{equation}
	\widehat{\mathbf{\Sigma}}(\mathbf{x}_0) = \sum_{ \mathbf{x_i} \in \mathcal{W}} w_i \mathbf{\Sigma(\mathbf{x}_i)}.
	\label{eqn:bfmean}
\end{equation}
 
Then the weights $w_i$ given by equation \eqref{eqn:bilat_weights} have to be expressed for matrices. Instead of norms of vector differences, matrix distances have to be used so that the weights act like similarities that attain a maximum for identical pixels and tend to zero if the distance goes to infinity. The spatial term of the weights remains unchanged.

 The new expression for the weights by replacing the Euclidean norm by a matrix distance $d(.,.)$ is
\begin{equation}
  w_i( \mathbf{x_i}) = \frac{  f_s(||\mathbf{x_i}-\mathbf{x_0}||_2)  
  f_r\left[ d(\mathbf{\Sigma}( \mathbf{x_i}) , \mathbf{\Sigma}(\mathbf{x_0}))\right] }
  {\sum_{ \mathbf{x_i} \in \mathcal{W}}  f_s(||\mathbf{x_i}-\mathbf{x_0}||_2)  
    f_r\left[ d(\mathbf{\Sigma}( \mathbf{x_i}) , \mathbf{\Sigma}(\mathbf{x_0}))\right]
  }.
  \label{eqn:bfweights}
\end{equation}

In the following we introduce several distances that are suitable for such a filter.

\subsection{The symmetrized Kullback-Leibler divergence}
\label{sub:MetricKL}
The Kullback-Leibler divergence \cite{Kullback1951} has been defined to characterize the discrepancy between two probability distributions $P_1$ and $P_2$
\begin{equation}
  D_{kl}(P_1\|P_2) = \int_{-\infty}^\infty p_1(x) \ln \frac{p_1(x)}{p_2(x)} \, {\rm d}x.
\end{equation}
Since this measure is not symmetric, a symmetrized Kullback-Leibler divergence may be defined as
\begin{equation}
  d_{kl}(P_1\|P_2) = \frac{1}{2} \left[D_{kl}(P_1\|P_2) + D_{kl}(P_2\|P_1)\right],
\end{equation}
and verifies $d_{kl}(P_1\|P_2) = d_{kl}(P_2\|P_1)$.
In the case of $d$-variate complex circular Normal distributions  $\mathcal{N}_1(0, \mathbf\Sigma_1)$ and $\mathcal{N}_2(0, \mathbf\Sigma_2)$, it can be shown that this pseudo-distance depends only on the covariance matrices and is expressed as
\begin{equation}
  d_{kl}(\mathbf{\Sigma_1},\mathbf{\Sigma_2}) = 
  \frac{1}{2}\mathrm{Tr} (\mathbf{\Sigma_1^{-1}}\mathbf{\Sigma_2} + \mathbf{\Sigma_2^{-1}}\mathbf{\Sigma_1}) - d,
\end{equation}
where $d$ is the dimension of the multivariate density.
Then, provided estimates of the covariances are available, this pseudo-distance can be used to compute the weights in equation \eqref{eqn:bfweights}.

\subsection{Distances on the cone of Hermitian positive definite matrices}
\label{sub:MetricsRiemann}
Since the covariance matrices lie on the symmetric cone of Hermitian positive definite (HPD) matrices, which is not a vector space but a Riemannian manifold, recent studies showed it was natural to use appropriate distances \cite{Pennec2006, Barbaresco2009}.
An affine invariant metric has been proposed as a replacement for the Euclidean metric to deal with HPD matrices. The corresponding distance between two matrices $\mathbf \Sigma_1$ and $\mathbf \Sigma_2$ is \cite{Barbaresco2009}
  \begin{equation}
	 d_{ai}(\mathbf{\Sigma_1},\mathbf{\Sigma_2}) = 
	 \lVert\log [\mathbf{\Sigma_1^{-\frac{1}{2}}} \mathbf{\Sigma_2}  \mathbf{\Sigma_1^{-\frac{1}{2}}} ] \rVert_F,
\end{equation}
where $\log()$ is the matrix logarithm and $\|.\|_F$ the Frobenius norm. This distance corresponds to the length of the geodesic (\textit{i.e.} the shortest path) between two points on the manifold and is invariant by affine transformation.

 Another choice of geodesics on the manifold leads to the simpler log-Euclidean distance that has been defined as \cite{Arsigny2006}
\begin{equation}
	d_{le}(\mathbf{\Sigma_1},\mathbf{\Sigma_2}) = \lVert \log(\mathbf{\Sigma_1}) - \log(\mathbf{\Sigma_2})\rVert_F.
\end{equation}
It has been shown that this distance is similarity invariant.
In the following we investigate the use of such distances to compute the weights in equation \eqref{eqn:bfweights}

\subsection{Iterative scheme and implementation details}
\label{sub:impl}
\paragraph{Iterative filtering} The distances introduced in the previous sections assume that the covariance matrices are known for each pixel of the PolSAR image. In practice, this is not the case since the original data is available in the form of the scattering matrix $\mathbf S$ described in equation \eqref{eqn:scatmatrix}. Moreover, those distances need to be computed on a full-rank matrix, that requires multi-looking  (defined in equation \eqref{eqn:multilooking}) of at least $d$ independent pixels. Therefore, this multi-look image, though leading to a noisy covariance estimate, will be used as an initialization $\mathbf{\widehat{\Sigma}}^{(0)}$ for filtering the image in an iterative way, so as to progressively refine the estimate \cite{Durand2002, Barash2002}
\begin{equation}
  \widehat{\mathbf{\Sigma}}^{(n+1)}(\mathbf{x}_0) = \sum_{ \mathbf{x_i} \in \mathcal{W}} w_i \mathbf{\widehat{\Sigma}}^{(n)}(\mathbf{x}_i),
	\label{eqn:bfiter}
\end{equation}
where the weights $w_i$ \eqref{eqn:bfweights} are computed on the covariances from iteration $n$.

\paragraph{Weight of the reference pixel} An issue of this type of filter when applied to noisy data is the very high importance given to the central pixel relative to its neighbours, leading to many unfiltered pixels. In fact, since this pixel is taken as a reference,  it is always assigned a weight of one (corresponding to a distance of zero). However, its value is uncertain because it is also affected by noise. Consequently, there is no reason to give this pixel such a high weight in the average. To avoid this excessive weighting, authors of \cite{Buades2011} propose to set the weight of the central pixel to the maximum weight among its neighbors. We use this recommendation in our experiments.

\paragraph{Rank 1 targets} 
The use of the statistical and geodesic distances involves matrix inversion and logarithm. In order to avoid numerical problems, the involved matrices have to be full-rank. In theory, in the case of Gaussian distributed targets, a minimal multi-looking ensures that those matrices have this property.  However, many pixels in PolSAR images cannot be described by distributed targets. If the resolution cell contains a dominant scatterer (due to, for instance, a corner reflector), the target behaviour is deterministic with one dominant scattering mechanism. The ideal model for this type of pixel is a rank 1 matrix, that cannot in theory be handled by the BLF. In practice, due to multi-looking and the scattering complexity of natural and made man targets, purely deterministic targets are unlikely to be present in the data. However, it is not excluded that some matrices are close to pure targets and produce numerical instability in the filtering process. For this reason, we propose a modification of the filter that allows to avoid the filtering of ill-conditioned matrices. 
For each pixel, before applying the filter, we compute the condition number of the matrix:
\begin{equation}
  \kappa(\mathbf{T}) = \frac{\lambda_{max}}{\lambda_{min}},
\end{equation}
where $\lambda_{max}$ and $\lambda_{min}$ are the maximal and minimal eigenvalues of the matrix. If the matrix is ill-conditioned, the condition number goes to infinity and the inverse of this value is close to zero \cite{Press1992}. Then we compare $1/\kappa(\mathbf{T})$ to a threshold (set to $10^{-6}$ in our experiments) to detect rank deficient matrices. If the pixel is the central pixel in the window, it is left unfiltered. If the pixel is not the central one, its weight is simply set to zero so that it is excluded from the local average.

\begin{figure}[t]
\centering
\includegraphics[scale=0.3]{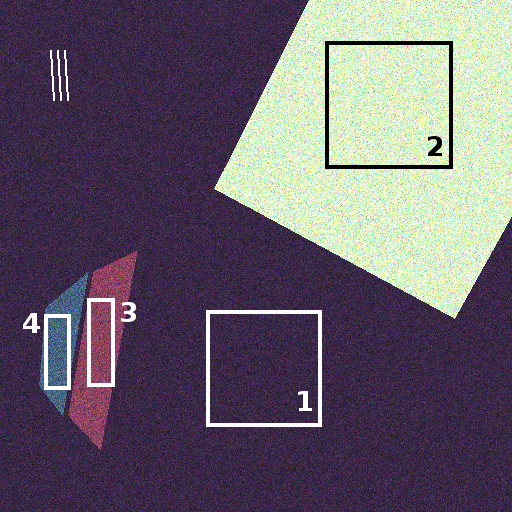}
\caption{$512 {\times} 512$ 4-look synthetic PolSAR image represented with Pauli RGB color coding. The rectangles represent the 4 selected homogeneous areas used for evaluation (see section \ref{sub:compare})}
\label{fig:simul}
\end{figure}

\begin{figure*}[t]
\centering
\subfigure[]{
\includegraphics[scale=0.45, trim=0.4cm 0.4cm 1.0cm 0.4cm, clip=true]{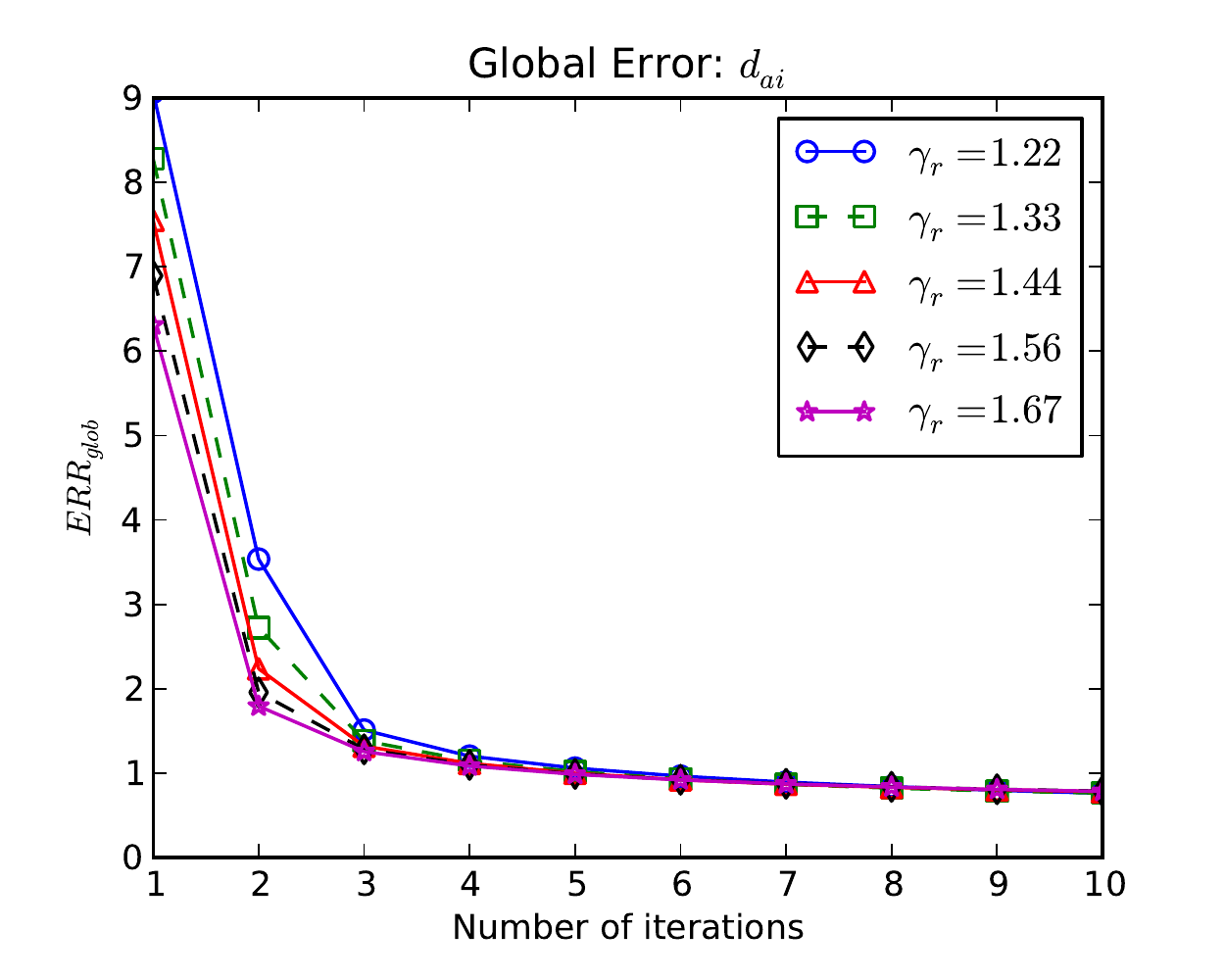}
\label{ca}
}
\subfigure[]{
\includegraphics[scale=0.45, trim=0.4cm 0.4cm 1.0cm 0.4cm, clip=true]{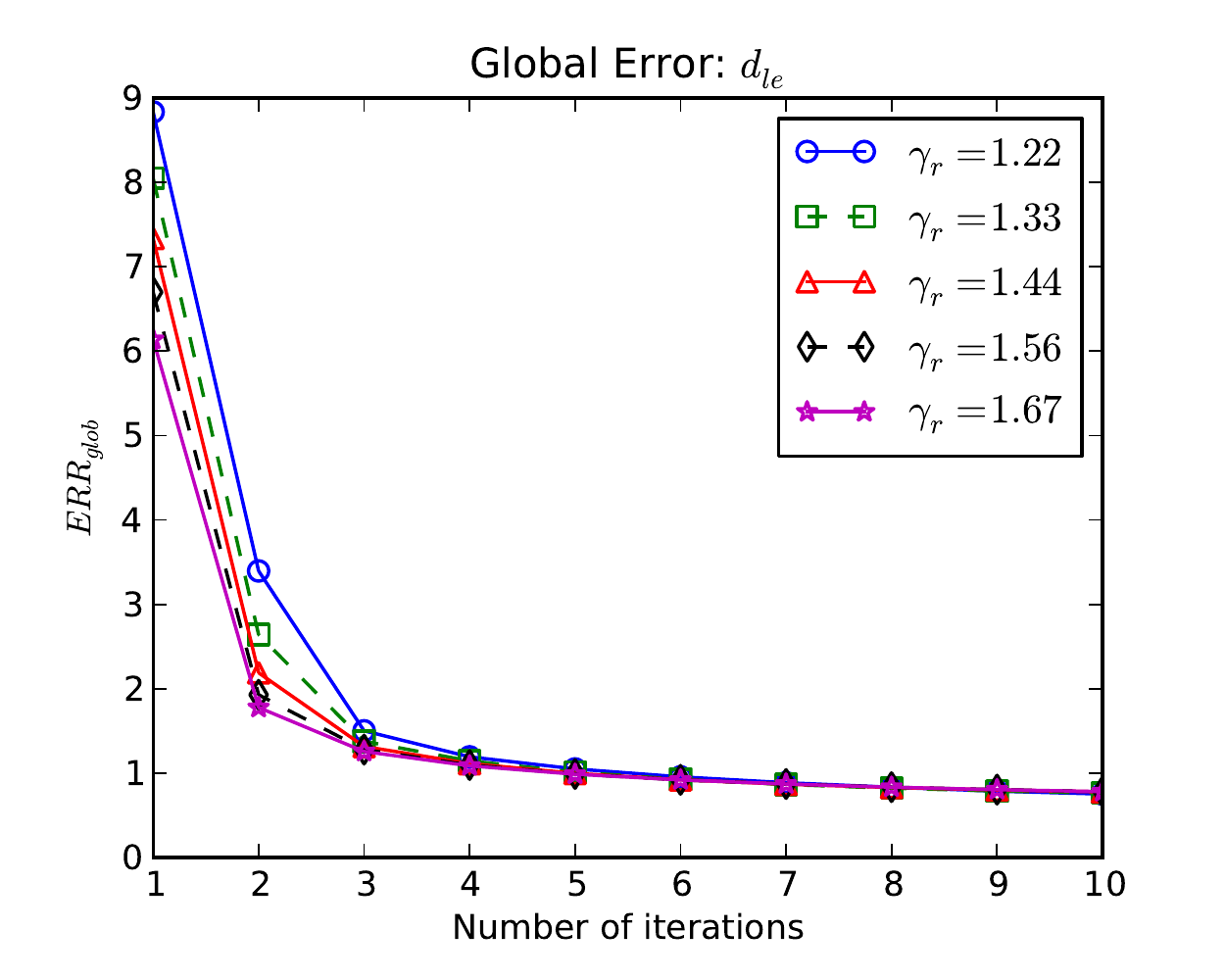}
\label{cb}
}
\subfigure[]{
\includegraphics[scale=0.45, trim=0.4cm 0.4cm 1.0cm 0.4cm, clip=true]{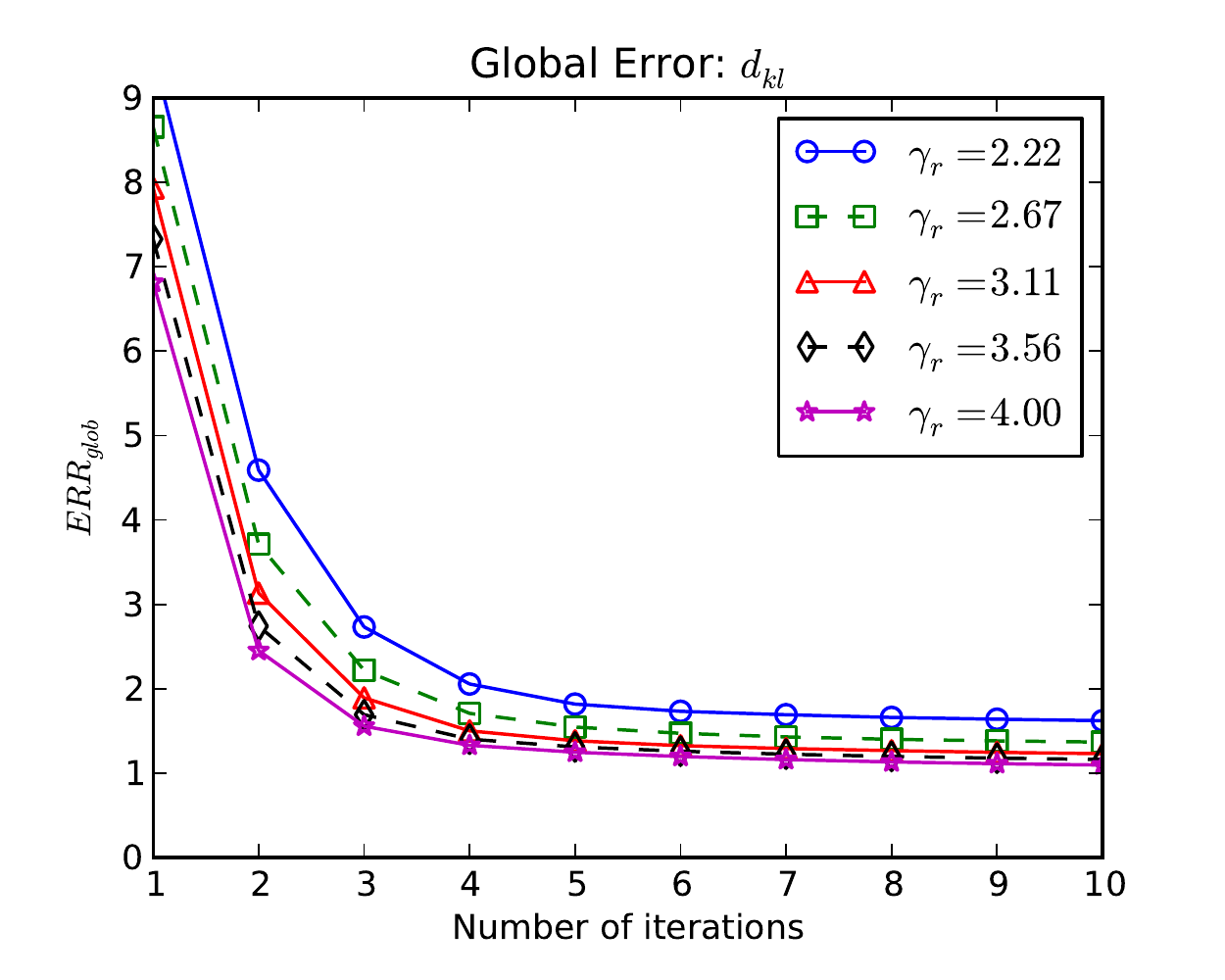}
\label{cc}
}\\
\subfigure[]{
\includegraphics[scale=0.45, trim=0.4cm 0.4cm 1.0cm 0.4cm, clip=true]{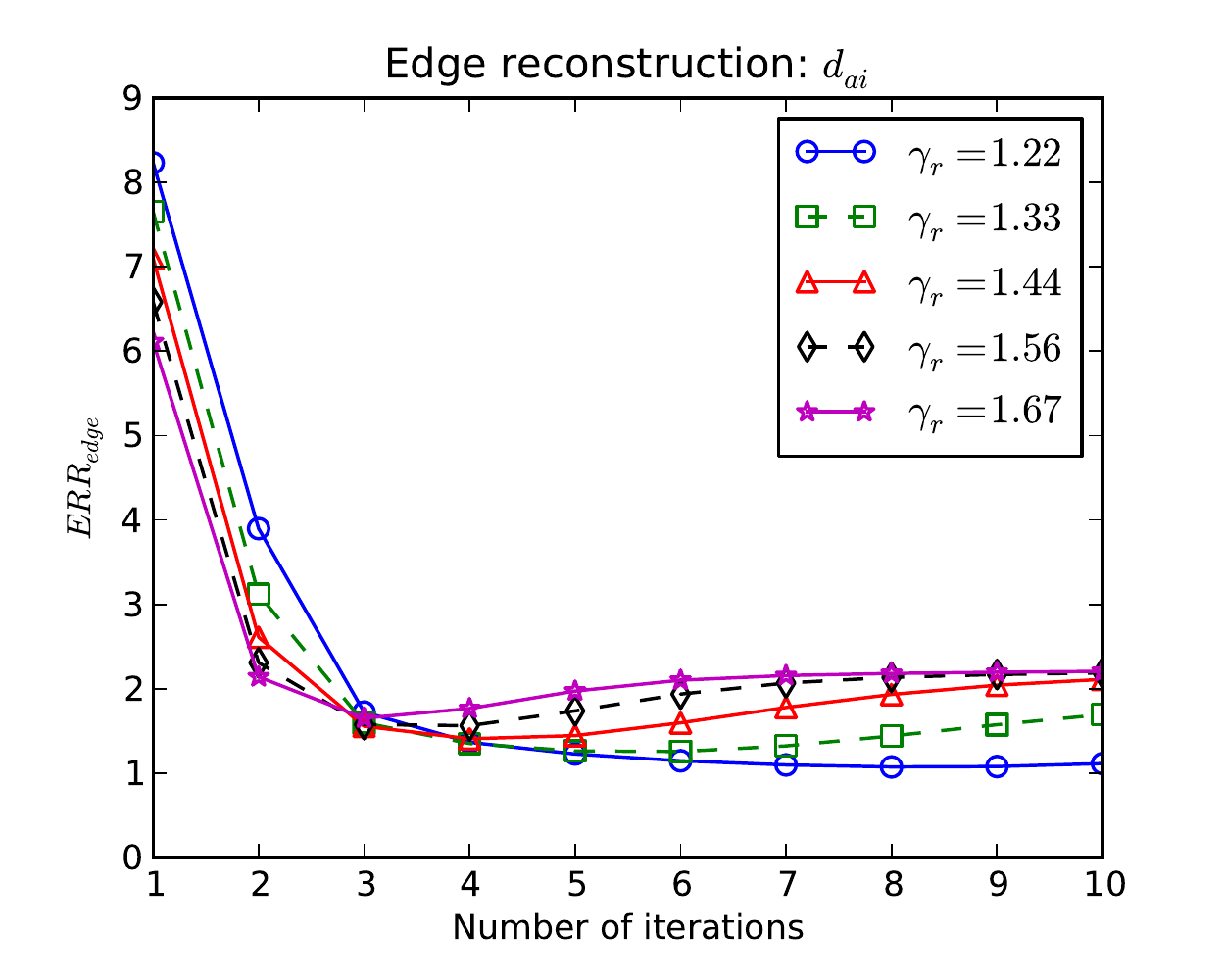}
\label{cd}
}
\subfigure[]{
\includegraphics[scale=0.45, trim=0.4cm 0.4cm 1.0cm 0.4cm, clip=true]{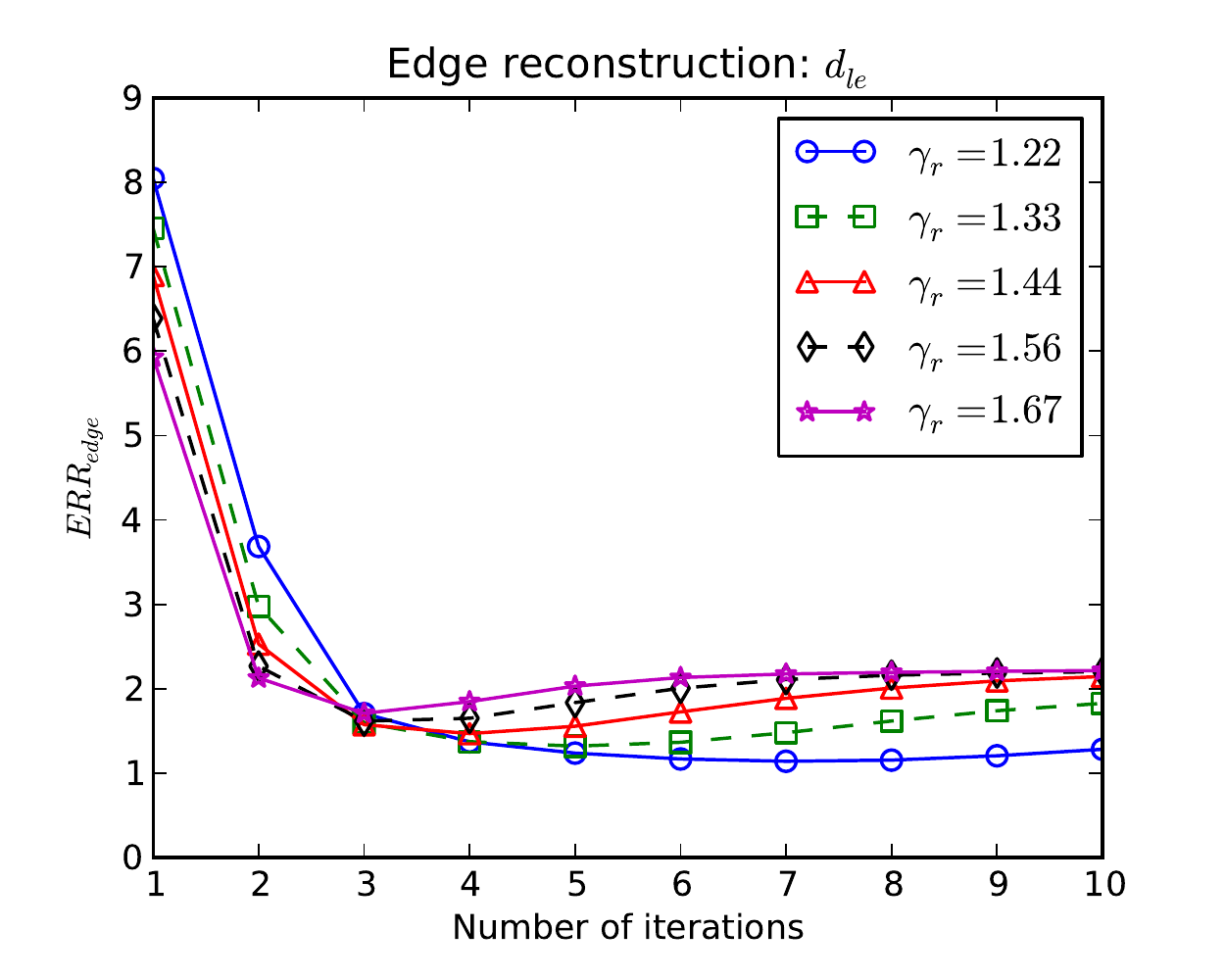}
\label{ce}
}
\subfigure[]{
\includegraphics[scale=0.45, trim=0.4cm 0.4cm 1.0cm 0.4cm, clip=true]{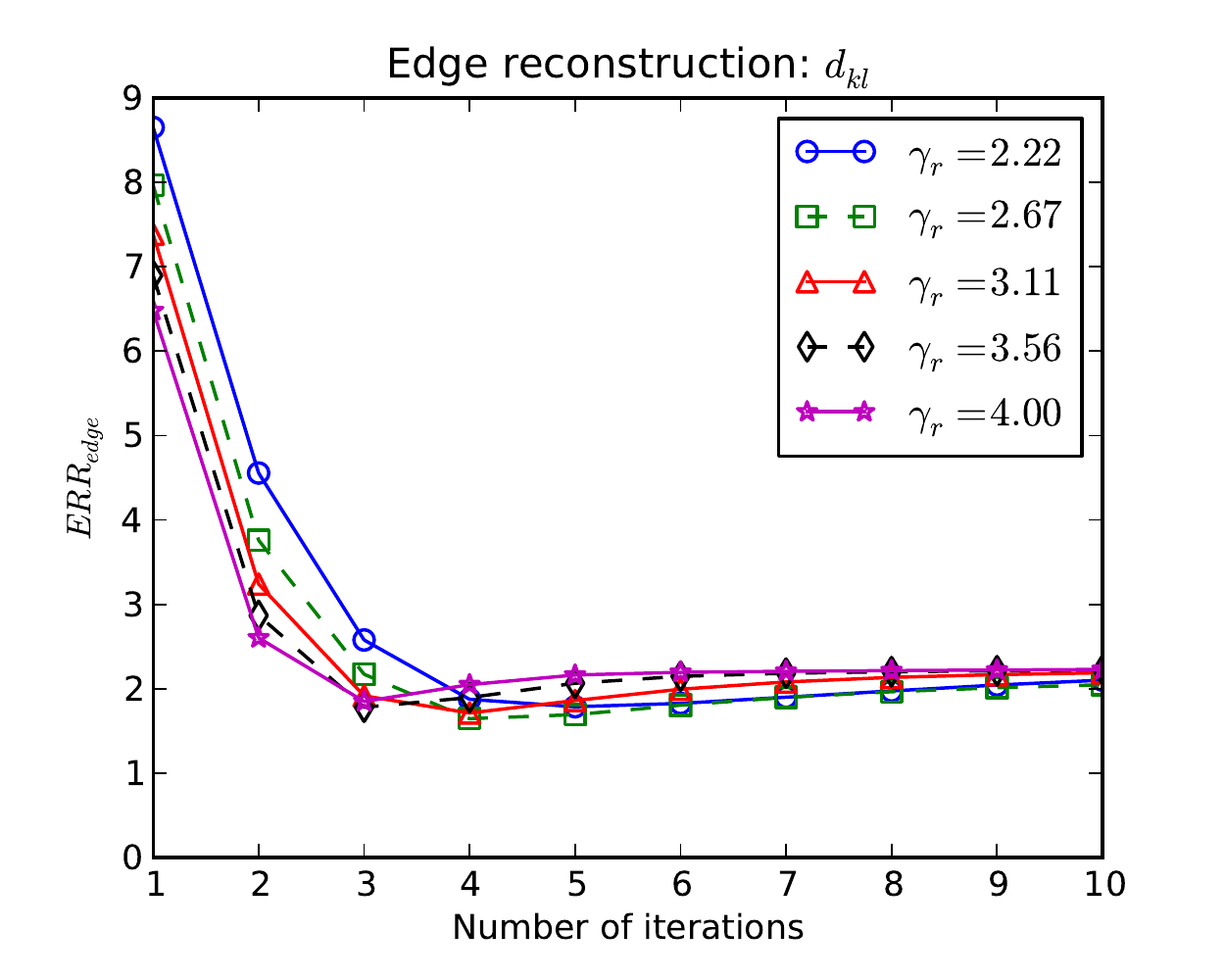}
\label{cf}
}\\
\subfigure[]{
\includegraphics[scale=0.44, trim=0.1cm 0.4cm 1.0cm 0.4cm, clip=true]{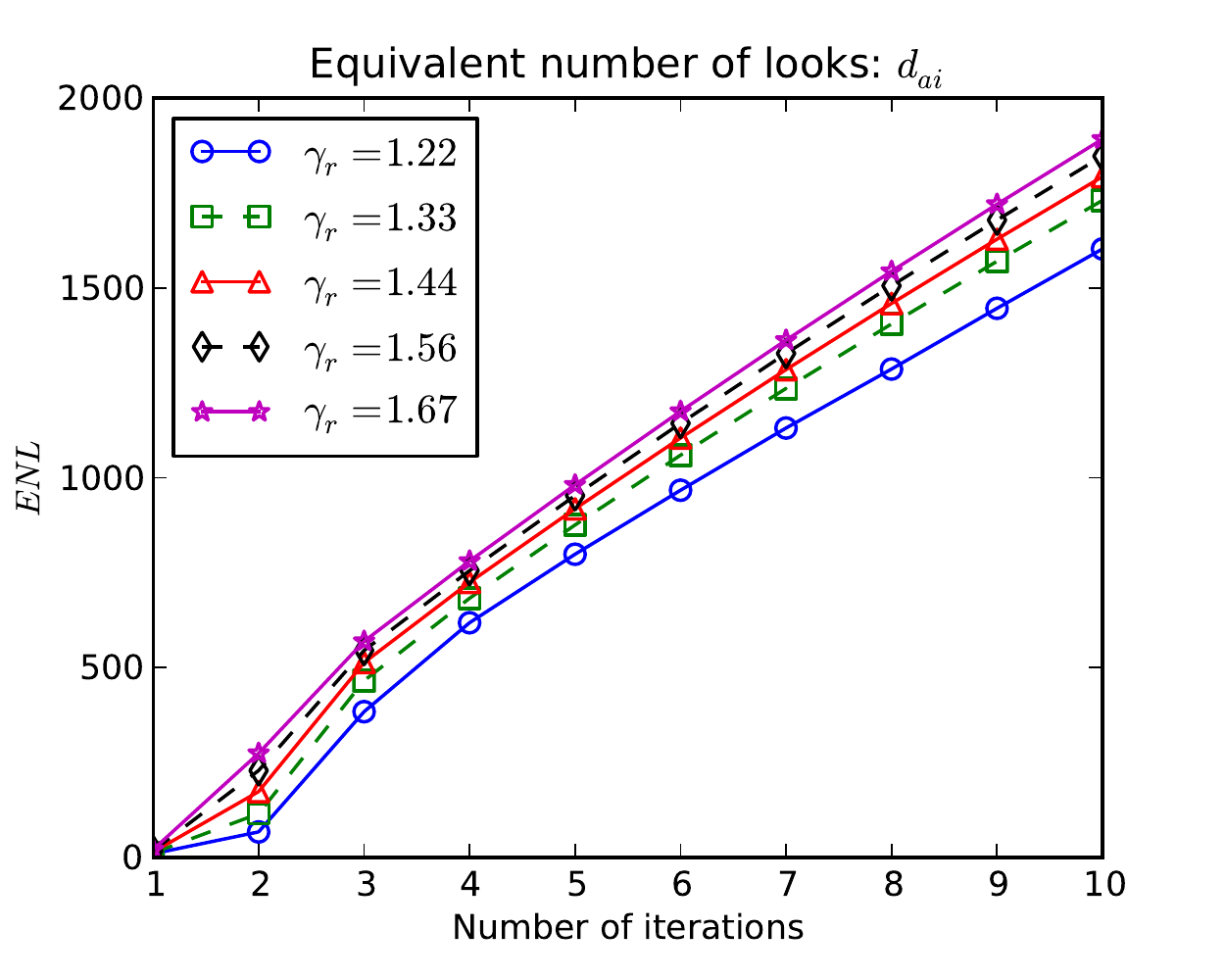}
\label{cg}
}
\subfigure[]{
\includegraphics[scale=0.44, trim=0.1cm 0.4cm 1.0cm 0.4cm, clip=true]{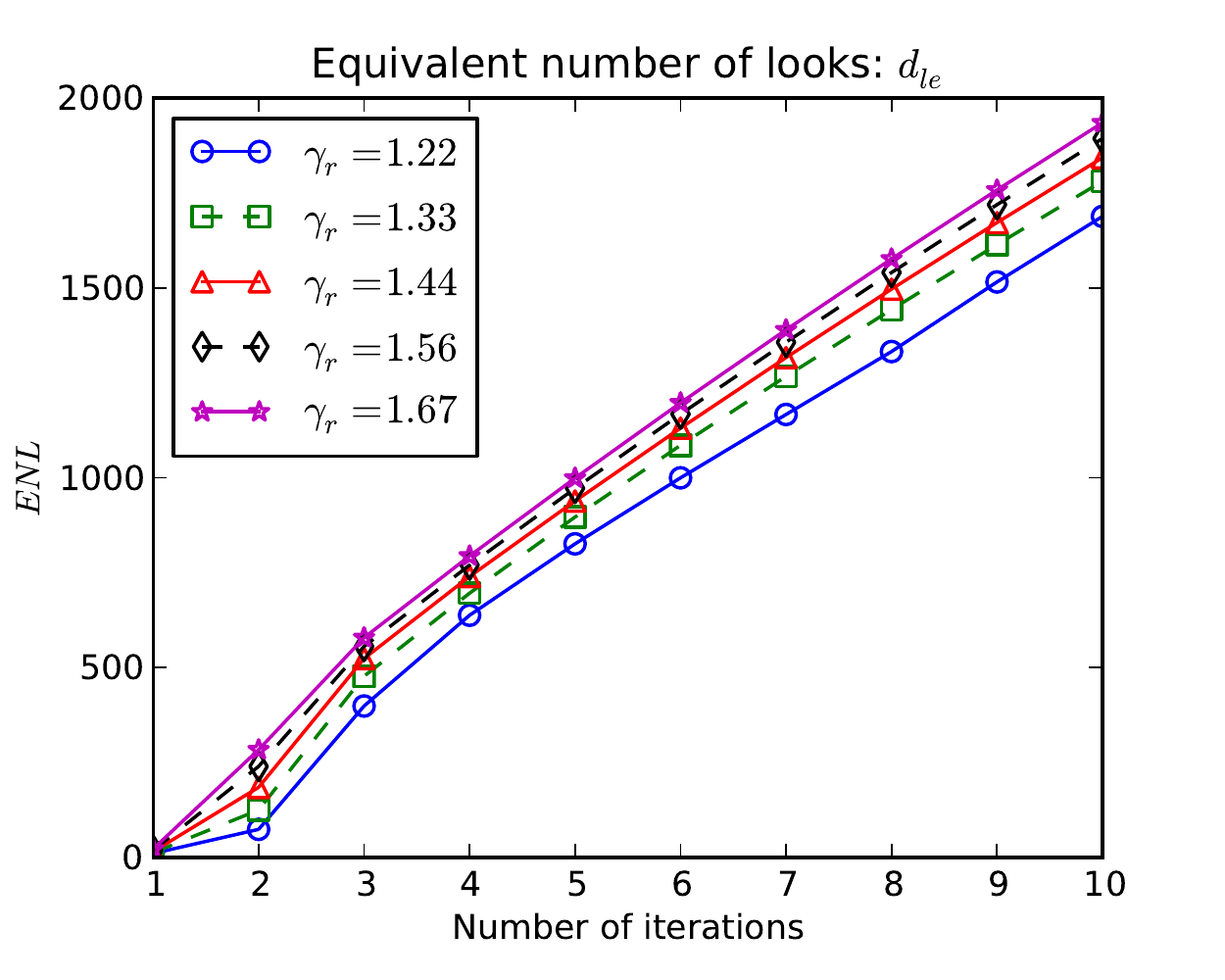}
\label{ch}
}
\subfigure[]{
\includegraphics[scale=0.44, trim=0.1cm 0.4cm 1.0cm 0.4cm, clip=true]{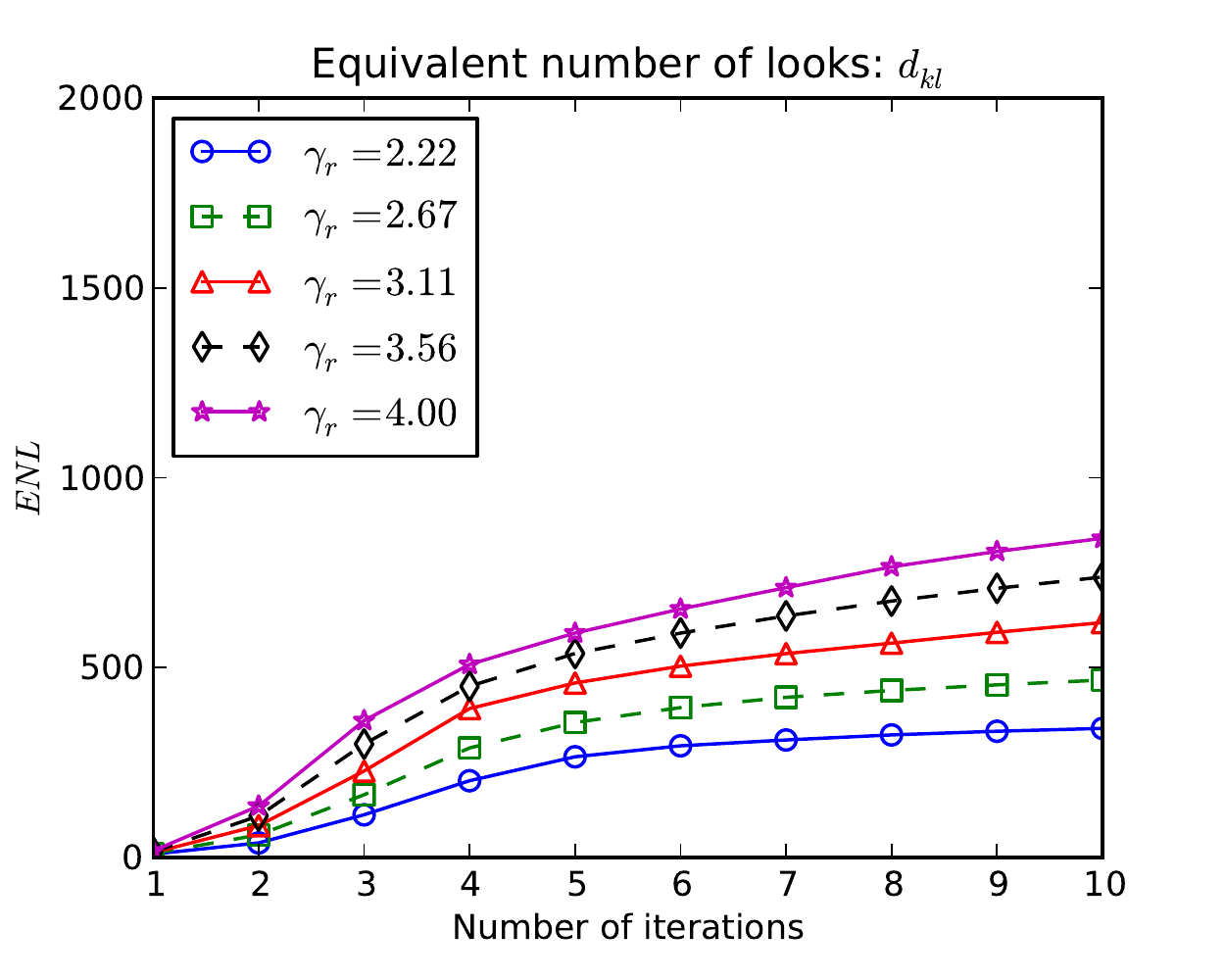}
\label{ci}
}
\caption{Performance of the different versions of the BLF filter: $d_{ai}$ (left column), $d_{le}$ (middle column), $d_{kl}$ (right column). The top line shows the global error $ERR_{glob}$, the middle line shows the edge reconstruction error $ERR_{egde}$ and the bottom line shows the equivalent number of looks $ENL$ (see text for details).}
\label{fig:curves}
\end{figure*}

\begin{figure}[t]
\centering
\subfigure[Original (cropped)]{
 \begin{minipage}[c][3.5cm]{1.75cm}
 \includegraphics[height=1.69cm, viewport=20 396 100 476, clip=true]{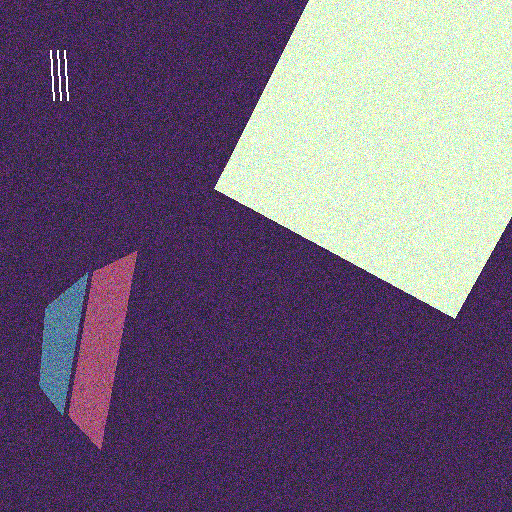}\\
\includegraphics[height=1.69cm, viewport=36 174 146 284, clip=true]{dhond4a} 
 \end{minipage}
 \begin{minipage}[c][3.5cm]{1.75cm}
\includegraphics[height=3.4cm, viewport=180 250 280 450, clip=true]{dhond4a}
 \end{minipage}
 \label{sfg:inflpara}
}
\subfigure[BLF $d_{ai}, \gamma_r {=} 1.33, N_{it} {=} 4$]{
 \begin{minipage}[c][3.5cm]{1.75cm}
 \includegraphics[height=1.69cm, viewport=20 396 100 476, clip=true]{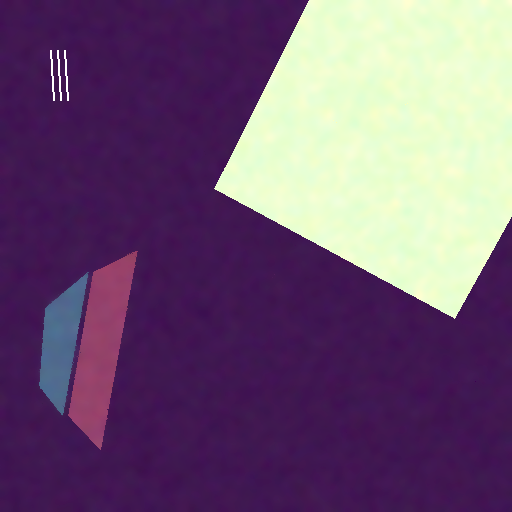}\\
\includegraphics[height=1.69cm, viewport=36 174 146 284, clip=true]{dhond4b} 
 \end{minipage}
 \begin{minipage}[c][3.5cm]{1.75cm}
\includegraphics[height=3.4cm, viewport=180 250 280 450, clip=true]{dhond4b}
 \end{minipage}
 \label{sfg:inflparb}
}\\
\subfigure[BLF $d_{ai}, \gamma_r {=} 1.33, N_{it} {=} 1$]{
 \begin{minipage}[c][3.5cm]{1.75cm}
 \includegraphics[height=1.69cm, viewport=20 396 100 476, clip=true]{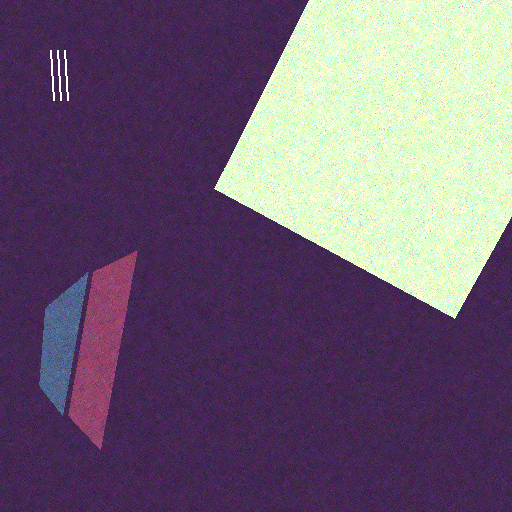}\\
\includegraphics[height=1.69cm, viewport=36 174 146 284, clip=true]{dhond4c} 
 \end{minipage}
 \begin{minipage}[c][3.5cm]{1.75cm}
\includegraphics[height=3.4cm, viewport=180 250 280 450, clip=true]{dhond4c}
 \end{minipage}
 \label{sfg:inflparc}
}
\subfigure[BLF $d_{ai}, \gamma_r {=} 1.33, N_{it} {=} 7$]{
 \begin{minipage}[c][3.5cm]{1.75cm}
 \includegraphics[height=1.69cm, viewport=20 396 100 476, clip=true]{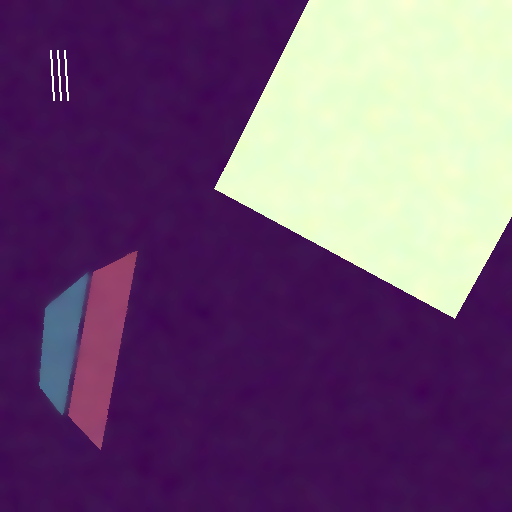}\\
\includegraphics[height=1.69cm, viewport=36 174 146 284, clip=true]{dhond4d} 
 \end{minipage}
 \begin{minipage}[c][3.5cm]{1.75cm}
\includegraphics[height=3.4cm, viewport=180 250 280 450, clip=true]{dhond4d}
 \end{minipage}
 \label{sfg:inflpard}
}\\
\subfigure[BLF $d_{ai}, \gamma_r {=} 0.7, N_{it} {=} 4$]{
 \begin{minipage}[c][3.5cm]{1.75cm}
 \includegraphics[height=1.69cm, viewport=20 396 100 476, clip=true]{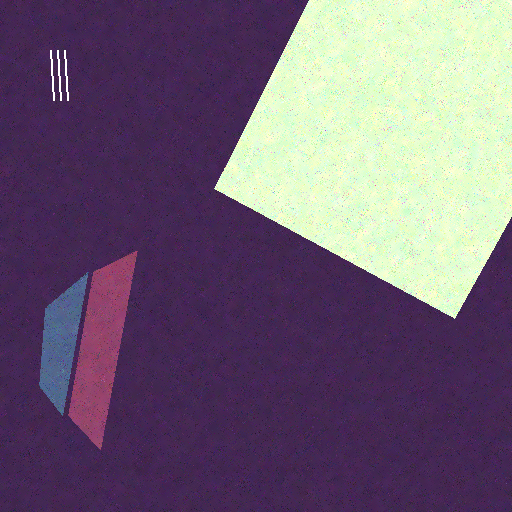}\\
\includegraphics[height=1.69cm, viewport=36 174 146 284, clip=true]{dhond4e} 
 \end{minipage}
 \begin{minipage}[c][3.5cm]{1.75cm}
\includegraphics[height=3.4cm, viewport=180 250 280 450, clip=true]{dhond4e}
 \end{minipage}
 \label{sfg:inflpare}
}
\subfigure[BLF $d_{ai}, \gamma_r {=} 2.0, N_{it} {=} 4$]{
 \begin{minipage}[c][3.5cm]{1.75cm}
 \includegraphics[height=1.69cm, viewport=20 396 100 476, clip=true]{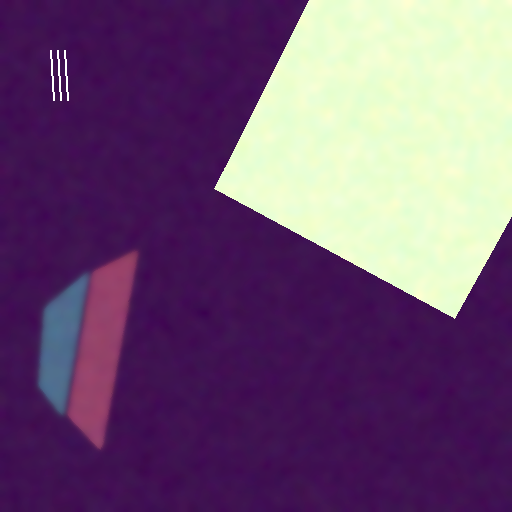}\\
\includegraphics[height=1.69cm, viewport=36 174 146 284, clip=true]{dhond4f} 
 \end{minipage}
 \begin{minipage}[c][3.5cm]{1.75cm}
\includegraphics[height=3.4cm, viewport=180 250 280 450, clip=true]{dhond4f}
 \end{minipage}
 \label{sfg:inflparf}
}

\caption{Influence of filter parameters on edge preservation and smoothing for the BLF with $d_{ai}$. For visualization purposes, we zoom on structures of interest. Those results can be interpreted in terms of edge preservation and noise reduction thanks to the curves of Fig. \ref{fig:curves}. The parameters used in \subref{sfg:inflparb} are those suggested by our quantitative analysis. Other choices of parameters lead to noisier \subref{sfg:inflparc}, \subref{sfg:inflpare}, or more blurred \subref{sfg:inflpard}, \subref{sfg:inflparf} results.}
\label{fig:paraminfl}
\end{figure}

\begin{figure*}[t]
\centering
\subfigure[Simulation]{
\includegraphics[scale=2, trim=0.6cm 0.6cm 0.6cm 0.6cm, clip=true]{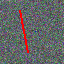}
\label{sfg:rank1a}
}
\subfigure[Filtered image]{
\includegraphics[scale=2, trim=0.6cm 0.6cm 0.6cm 0.6cm, clip=true]{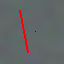}
\label{sfg:rank1b}
}
\subfigure[$H$ (true value)]{
\includegraphics[scale=2, trim=0.6cm 0.6cm 0.6cm 0.6cm, clip=true]{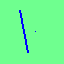}
\label{sfg:rank1c}
}
\subfigure[$\alpha$ (true value)]{
\includegraphics[scale=2, trim=0.6cm 0.6cm 0.6cm 0.6cm, clip=true]{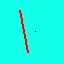}
\label{sfg:rank1d}
}
\subfigure[$H$ (filtered)]{
\includegraphics[scale=2, trim=0.6cm 0.6cm 0.6cm 0.6cm, clip=true]{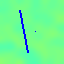}
\label{sfg:rank1e}
}
\subfigure[$\alpha$ (filtered)]{
\includegraphics[scale=2, trim=0.6cm 0.6cm 0.6cm 0.6cm, clip=true]{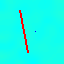}
\label{sfg:rank1f}
}
\subfigure{
  \raisebox{-1mm}{\includegraphics[scale=0.33]{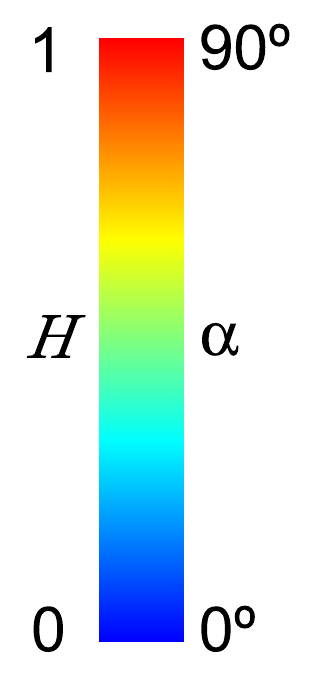}}
}

\caption{Evaluation on rank 1 targets. Two deterministic targets are artificially generated by simulating a pure trihedral coherency matrix (blue dot) and a pure dihedral one (red line). The background is speckle simulated by the Monte Carlo method described in section \ref{sscn:SimVal}. The simulated data \subref{sfg:rank1a} is filtered with the BLF $d_{ai}$ (parameters are chosen according to section \ref{sscn:SimVal}) \subref{sfg:rank1b} and we observe that the true $H/\alpha$ parameters \subref{sfg:rank1c},  \subref{sfg:rank1d} are similar to the filtered ones \subref{sfg:rank1e}, \subref{sfg:rank1f} . }
\label{fig:rank1}
\end{figure*}

\begin{figure}[t]
\centering
\subfigure[Boxcar $7\times7$]{
 \begin{minipage}[c][3.5cm]{1.75cm}
 \includegraphics[height=1.69cm, viewport=20 396 100 476, clip=true]{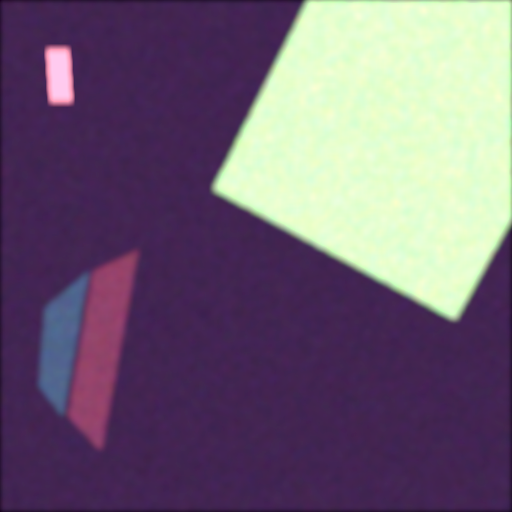}\\
\includegraphics[height=1.69cm, viewport=36 174 146 284, clip=true]{dhond6a} 
 \end{minipage}
 \begin{minipage}[c][3.5cm]{1.75cm}
\includegraphics[height=3.4cm, viewport=180 250 280 450, clip=true]{dhond6a}
 \end{minipage}
 \label{sfg:simfilta}
}
\subfigure[BLF $d_{kl}$, $\gamma_r {=} 3.11$, $N_{it} {=} 4$]{
 \begin{minipage}[c][3.5cm]{1.75cm}
 \includegraphics[height=1.69cm, viewport=20 396 100 476, clip=true]{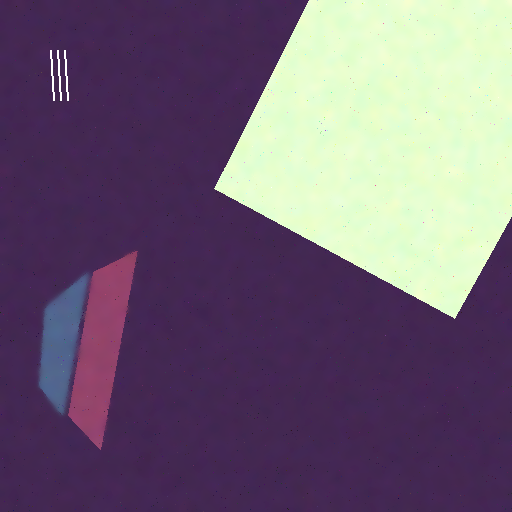}\\
\includegraphics[height=1.69cm, viewport=36 174 146 284, clip=true]{dhond6b} 
 \end{minipage}
 \begin{minipage}[c][3.5cm]{1.75cm}
\includegraphics[height=3.4cm, viewport=180 250 280 450, clip=true]{dhond6b}
 \end{minipage}
 \label{sfg:simfiltb}
}\\
\subfigure[BLF $d_{le}$, $\gamma_r {=} 1.33$, $N_{it} {=} 4$]{
 \begin{minipage}[c][3.5cm]{1.75cm}
 \includegraphics[height=1.69cm, viewport=20 396 100 476, clip=true]{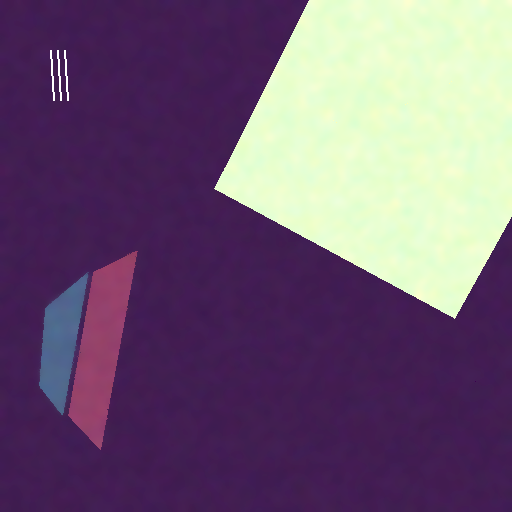}\\
\includegraphics[height=1.69cm, viewport=36 174 146 284, clip=true]{dhond6c} 
 \end{minipage}
 \begin{minipage}[c][3.5cm]{1.75cm}
\includegraphics[height=3.4cm, viewport=180 250 280 450, clip=true]{dhond6c}
 \end{minipage}
 \label{sfg:simfiltc}
}
\subfigure[BLF $d_{ai}$, $\gamma_r {=} 1.33$, $N_{it} {=} 4$]{
 \begin{minipage}[c][3.5cm]{1.75cm}
 \includegraphics[height=1.69cm, viewport=20 396 100 476, clip=true]{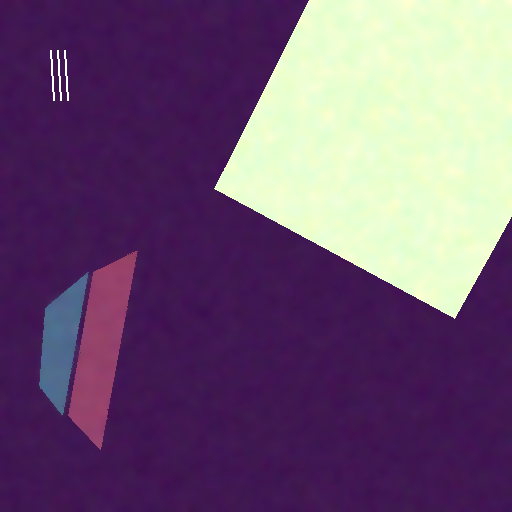}\\
\includegraphics[height=1.69cm, viewport=36 174 146 284, clip=true]{dhond6d} 
 \end{minipage}
 \begin{minipage}[c][3.5cm]{1.75cm}
\includegraphics[height=3.4cm, viewport=180 250 280 450, clip=true]{dhond6d}
 \end{minipage}
 \label{sfg:simfiltd}
}\\
\subfigure[Refined Lee]{
 \begin{minipage}[c][3.5cm]{1.75cm}
 \includegraphics[height=1.69cm, viewport=20 396 100 476, clip=true]{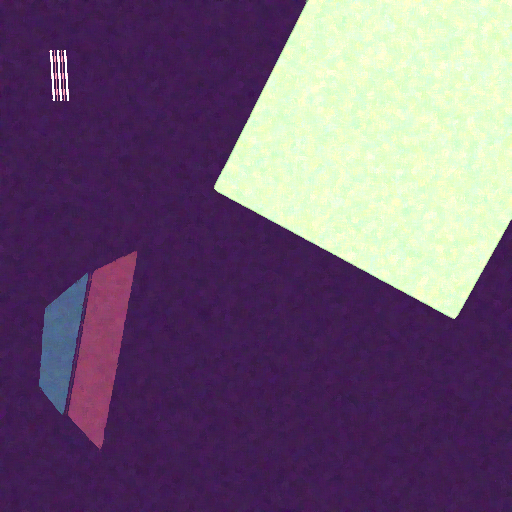}\\
\includegraphics[height=1.69cm, viewport=36 174 146 284, clip=true]{dhond6e} 
 \end{minipage}
 \begin{minipage}[c][3.5cm]{1.75cm}
\includegraphics[height=3.4cm, viewport=180 250 280 450, clip=true]{dhond6e}
 \end{minipage}
 \label{sfg:simfilte}
}
\subfigure[IDAN]{
 \begin{minipage}[c][3.5cm]{1.75cm}
 \includegraphics[height=1.69cm, viewport=20 396 100 476, clip=true]{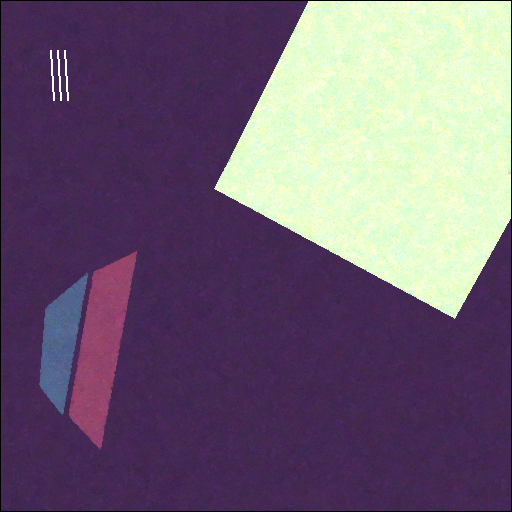}\\
\includegraphics[height=1.69cm, viewport=36 174 146 284, clip=true]{dhond6f} 
 \end{minipage}
 \begin{minipage}[c][3.5cm]{1.75cm}
\includegraphics[height=3.4cm, viewport=180 250 280 450, clip=true]{dhond6f}
 \end{minipage}
 \label{sfg:simfiltf}
}
\caption{Comparison of the different filtering techniques over the synthetic image. The image is cropped for a better visualization. The boxcar filter \subref{sfg:simfilta} blurs edges while the different versions of the BLF \subref{sfg:simfiltb}, \subref{sfg:simfiltc}, \subref{sfg:simfiltd} allow a high level of smoothing with better edge preservation. Refined Lee \subref{sfg:simfilte} and IDAN \subref{sfg:simfiltf} results are shown for comparison.}
\label{fig:simfilt}
\end{figure}

\begin{table*}[t]
\caption{Performance comparison for different filters (see text for details).}
\label{fig:comperr}
\centering
\begin{tabular}{|c||c|c|c|c|c|c|}
\hline
	& Box & Ref. Lee & IDAN & BLF $d_{kl}$ & BLF $d_{ai}$ & BLF $d_{le}$\\
	\hline	\hline
	$ERR_{glob}$ & 6.83 & 3.43 &  2.63 & 1.50 & 1.15 & 1.14 \\
	\hline
	$ERR_{edge}$ & 54.5 & 17.3 & 6.90 & 1.71 & 1.35 & 1.37 \\
	\hline
	$ENL$ & 206 & 95.3 & 86.3 & 492 & 683 & 696 \\
	\hline
\end{tabular}
\end{table*}


\begin{table*}[t]
\centering \footnotesize
\caption{Comparison of mean estimates of entropy, mean alpha angle and $\mathbf T$ matrix elements over four homogeneous areas for the considered PolSAR speckle reduction methods (see text for details).}
\label{fig:compcoh}
\subfigure[Zone 1]{
\begin{tabular}{|c||c|c|c|c|c|c|c|}
\hline
  & True value  & Box $7\times7$  & Ref. Lee  & IDAN  & BF $d_{kl}$  & BF $d_{ai}$  & BF $d_{le}$ \\
\hline\hline
$H$ & 0.48  & 0.48  & 0.48  & 0.49  & 0.48  & 0.48  & 0.47 \\
\hline
$\alpha$ & 0.56  & 0.56  & 0.56  & 0.56  & 0.56  & 0.56  & 0.56 \\
\hline
$T_{11}$ & 8.03  & 8.04  & 7.99  & 7.46  & 7.90  & 7.92  & 8.12 \\
\hline
$T_{22}$ & 2.64  & 2.63  & 2.63  & 2.44  & 2.59  & 2.60  & 2.64 \\
\hline
$T_{33}$ & 0.55  & 0.55  & 0.55  & 0.51  & 0.54  & 0.54  & 0.53 \\
\hline
$T_{12}$ & -2.19 - j 2.23  & -2.19 - j 2.23  & -2.17 - j 2.21  & -1.98 - j 2.01  & -2.15 - j 2.19  & -2.16 - j 2.20  & -2.22 - j 2.26 \\
\hline
$T_{13}$ & -0.17 - j 0.15  & -0.17 - j 0.15  & -0.16 - j 0.15  & -0.14 - j 0.13  & -0.16 - j 0.15  & -0.16 - j 0.15  & -0.17 - j 0.15 \\
\hline
$T_{23}$ & 0.11 - j 0.03  & 0.11 - j 0.04  & 0.11 - j 0.04  & 0.09 - j 0.03  & 0.11 - j 0.04  & 0.11 - j 0.04  & 0.11 - j 0.04 \\
\hline
\end{tabular}
}

\subfigure[Zone 2]{
\begin{tabular}{|c||c|c|c|c|c|c|c|}
\hline
  & True value  & Box $7\times7$  & Ref. Lee  & IDAN  & BF $d_{kl}$  & BF $d_{ai}$  & BF $d_{le}$ \\
\hline\hline
$H$ & 0.97  & 0.97  & 0.96  & 0.97  & 0.97  & 0.97  & 0.97 \\
\hline
$\alpha$ & 0.87  & 0.88  & 0.88  & 0.88  & 0.88  & 0.88  & 0.88 \\
\hline
$T_{11}$ & 75.21  & 74.95  & 74.83  & 70.61  & 73.64  & 73.93  & 74.57 \\
\hline
$T_{22}$ & 48.03  & 47.92  & 48.15  & 45.23  & 47.16  & 47.34  & 47.32 \\
\hline
$T_{33}$ & 45.82  & 46.04  & 46.30  & 43.72  & 45.44  & 45.58  & 45.50 \\
\hline
$T_{12}$ & 4.86 + j 3.24  & 4.72 + j 2.95  & 4.59 + j 2.87  & 4.07 + j 2.58  & 4.57 + j 2.85  & 4.55 + j 2.85  & 4.68 + j 2.93 \\
\hline
$T_{13}$ & 2.30 + j 0.22  & 2.33 + j 0.56  & 2.28 + j 0.54  & 2.17 + j 0.57  & 2.33 + j 0.58  & 2.32 + j 0.56  & 2.37 + j 0.59 \\
\hline
$T_{23}$ & -0.32 - j 1.69  & -0.42 - j 1.77  & -0.37 - j 1.73  & -0.57 - j 1.46  & -0.40 - j 1.70  & -0.41 - j 1.69  & -0.42 - j 1.73 \\
\hline
\end{tabular}
}

\subfigure[Zone 3]{
\begin{tabular}{|c||c|c|c|c|c|c|c|}
\hline
  & True value  & Box $7\times7$  & Ref. Lee  & IDAN  & BF $d_{kl}$  & BF $d_{ai}$  & BF $d_{le}$ \\
\hline\hline
$H$ & 0.68  & 0.68  & 0.68  & 0.69  & 0.68  & 0.68  & 0.68 \\
\hline
$\alpha$ & 0.82  & 0.83  & 0.83  & 0.82  & 0.82  & 0.82  & 0.82 \\
\hline
$T_{11}$ & 13.71  & 13.32  & 13.24  & 12.61  & 13.23  & 13.31  & 13.51 \\
\hline
$T_{22}$ & 13.82  & 13.55  & 13.39  & 12.71  & 13.34  & 13.41  & 13.64 \\
\hline
$T_{33}$ & 1.55  & 1.55  & 1.56  & 1.46  & 1.52  & 1.52  & 1.50 \\
\hline
$T_{12}$ & 2.41 + j 5.86  & 2.44 + j 5.75  & 2.38 + j 5.66  & 2.19 + j 5.24  & 2.37 + j 5.72  & 2.37 + j 5.77  & 2.44 + j 5.92 \\
\hline
$T_{13}$ & -0.25 - j 0.29  & -0.26 - j 0.27  & -0.27 - j 0.29  & -0.23 - j 0.30  & -0.25 - j 0.26  & -0.26 - j 0.26  & -0.26 - j 0.26 \\
\hline
$T_{23}$ & 0.89 - j 0.16  & 0.92 - j 0.12  & 0.91 - j 0.13  & 0.85 - j 0.13  & 0.90 - j 0.12  & 0.90 - j 0.11  & 0.93 - j 0.11 \\
\hline\end{tabular}
}

\subfigure[Zone 4]{
\begin{tabular}{|c||c|c|c|c|c|c|c|}
\hline
  & True value  & Box $7\times7$  & Ref. Lee  & IDAN  & BF $d_{kl}$  & BF $d_{ai}$  & BF $d_{le}$ \\
\hline\hline
$H$ & 0.54  & 0.54  & 0.54  & 0.55  & 0.54  & 0.54  & 0.53 \\
\hline
$\alpha$ & 0.45  & 0.45  & 0.45  & 0.45  & 0.45  & 0.44  & 0.44 \\
\hline
$T_{11}$ & 25.71  & 25.45  & 25.18  & 23.66  & 25.04  & 25.22  & 25.80 \\
\hline
$T_{22}$ & 3.79  & 3.78  & 3.77  & 3.56  & 3.72  & 3.72  & 3.72 \\
\hline
$T_{33}$ & 3.40  & 3.38  & 3.39  & 3.17  & 3.31  & 3.32  & 3.31 \\
\hline
$T_{12}$ & 2.67 - j 3.48  & 2.55 - j 3.35  & 2.51 - j 3.28  & 2.22 - j 3.00  & 2.50 - j 3.31  & 2.52 - j 3.34  & 2.60 - j 3.42 \\
\hline
$T_{13}$ & -2.94 - j 1.56  & -2.83 - j 1.61  & -2.80 - j 1.61  & -2.49 - j 1.38  & -2.80 - j 1.55  & -2.81 - j 1.56  & -2.88 - j 1.62 \\
\hline
$T_{23}$ & -0.57 - j 0.86  & -0.61 - j 0.86  & -0.61 - j 0.85  & -0.53 - j 0.73  & -0.59 - j 0.84  & -0.59 - j 0.84  & -0.60 - j 0.86 \\
\hline
\end{tabular}
}

\end{table*}

\section{Experiments}
\label{sec:Experiments}
To validate the method, we have divided our experiments into several main steps: we first evaluate the influence of the filter parameters on simulated data generated from  sample covariances measured on an experimental dataset. This evaluation helps us to tune the filter parameters. Then, we compare our approach to other filters from the literature. Finally, we apply the method to experimentally acquired data sets.


\subsection{Validation on simulated data}
\label{sscn:SimVal}
Let us first introduce the strategy employed to generate our synthetic data. To obtain realistic simulations, we have selected four homogeneous areas from the fully polarimetric L-band Oberpfaffenhofen image acquired by the ESAR sensor from DLR in 1998. The mean $\mathbf T$ matrices over these areas have then been computed and used to simulate the 4 region dataset shown in Fig. \ref{fig:simul} according to the method described in \cite{Lee2009} (pp. 114-115) that we briefly recall here:
\begin{enumerate}
  \item Compute the square root $\mathbf{T}^{1/2}$ of the covariance to simulate $\mathbf{T}$, where $\mathbf{T}^{1/2}(\mathbf{T}^{1/2})^\dagger = \mathbf{T}$.
  \item Simulate $L$ complex random vectors $\mathbf{v_i}$ with zero mean and identity covariance matrix. 
  \item Compute $L$ single-look vectors $\mathbf{k}_i = \mathbf{T}^{1/2}\mathbf{v_i}$.
  \item Compute the multi-look covariance $\mathbf{\Sigma}$ by combining the independent samples according to equation \eqref{eqn:multilooking}.
\end{enumerate}
It may be noted that those simulations only allow to generate homogeneous areas according to the fully-developed speckle model. The case of non-Gaussian and spatially correlated data are beyond the scope of this paper and left for future work.

Additionally, to simulate deterministic targets, we have measured the coherency matrix over a building layover in the ESAR image and added it in the form of parallel lines (close to the top left corner of our synthetic image) without speckle. Assuming that the strong double bounce scattering from the building dominates the overall response, the effects of speckle can be neglected in this case. Note that this is only an approximate way to simulate this kind of target. Nevertheless, we consider it is sufficient to study the spatial properties of our filter in this simulation. In fact, purely deterministic targets are represented by rank 1 matrices (only one scattering mechanism is present). This ideal model is usually not found in experimental data due to the presence of speckle and the effect of multi-looking. However, it is important to evaluate the filter in such an extreme case since some targets may be close to rank 1. Therefore, we study this case in a separate simulation, that is described in section \ref{sscn:exprank1}.  

This simulated dataset allows a quantitative evaluation of the bilateral filter (that we refer to as BLF in the following) for the three distances $d_{kl}$, $d_{ai}$ and $d_{le}$ described in sections \ref{sub:MetricKL} and \ref{sub:MetricsRiemann}. The BLF has 3 parameters: $\gamma_s$, $\gamma_r$ and the number of iterations $N_{it}$. Since $\gamma_s$ has the role of a spatial window, which has already been extensively studied for speckle filtering \cite{Touzi2002}, we fix it to $2.2$ and use an $11\times 11$ window in all experiments. In the following, we focus our study on $\gamma_r$ and $N_{it}$.

We have chosen quantitative performance measures according to the following criteria:
\begin{enumerate}
	\item closeness to the original signal in the mean square sense,
	\item amount of smoothing in homogeneous areas and
        \item preservation of edges. 
\end{enumerate}
To evaluate the performance according to criterion 1, we propose the following quantity:
\begin{equation}
  ERR_{glob} = \left[\frac{1}{Nd^2} \sum_{i=1}^{N} \| \mathbf{\widehat{T}}_{i} - \mathbf{T}_{i} \|_F^2\right]^{1/2},
  \label{eqn:errglob}
\end{equation}
where $N$ is the number of pixels in the image, $d$ is the dimension of the matrices, $\mathbf T_i$ is the true covariance at pixel $i$ and $\mathbf{\widehat{T}}_i$ its estimate by the BLF.
This error is the per element root mean square reconstruction error, measuring the average deviation between the true covariances and the ones estimated by the filter.

To measure the amount of smoothing over homogeneous areas (criterion 2), we use the estimated equivalent number of looks (ENL) \cite{Lee2009}
\begin{equation}
  ENL = \frac{\widehat\mu^2}{\widehat\sigma^2},
\end{equation}
where $\widehat\mu$ and $\widehat\sigma^2$ are the estimated mean and variance over a considered channel. 
Since our filter performs averages of covariance matrices and does not process the elements independently, all the channels are processed identically. Consequently, the ENL is only measured on the $T_{11}$ channel of the filtered image and over a manually selected homogeneous area, assuming the other channels have been applied the same smoothing.

Finally, an important feature of speckle filters that is sometimes overlooked is the preservation of edges (criterion 3). Here, an edge is considered to be present if two adjacent pixels belong to a different class in the ground-truth used for simulations. It is difficult to evaluate edge preservation on PolSAR images due to the complex nature of the data. We have found experimentally that the measure $ERR_{glob}$ as defined previously was not a relevant measure of edge reconstruction.
This is due to the fact that pixels from homogeneous areas 
largely outnumber the ones located around edges and tend to dominate this quantity. Some authors \cite{Yu2002, Foucher2009} propose to use gradients to evaluate edge preservation.
However, gradients can only be computed on intensity and are highly sensitive to noise.
To measure the polarimetric reconstruction of edges, we propose to use the same measure as in equation \eqref{eqn:errglob} restricted to the pixels that are located on both sides of discontinuities
\begin{equation}
  ERR_{edge} = \left[\frac{1}{d^2} 
    \frac{\sum_{i=1}^{N} \delta(M_i\neq 0) \| \mathbf{\widehat{T}}_{i} - \mathbf{T}_{i} \|_F^2}
    {\sum_{i=1}^{N} \delta(M_i\neq 0)} 
  \right]^{1/2},
  \label{eqn:erredge}
\end{equation}
where $M_i$ is the binary mask of such pixels indexed by their location $i$, $d$ is defined as in eq. \eqref{eqn:errglob} and $\delta(.)$ is equal to $1$ if its argument is true and $0$ otherwise.
The mask $M_i$ is set equal to one if the pixel at location $i$ has at least one neighbour belonging to a different class in the ground truth.
Thus, we eliminate the influence of pixels located inside homogeneous areas and focus on edge reconstruction.

Fig. \ref{fig:curves} represents the curves corresponding to the variation of those values for different values of BLF parameters. 
We chose the range of values experimentally considering the fact that the curves for $ERR_{edge}$ have a local minimum for certain combinations of $\gamma_r$ and $N_{it}$. We limited the number of displayed curves for visual purpose. However, the evolution of the curves allow the reader to understand the effect of those parameters on the different measures. On the one hand, for low values of $\gamma_r$ the local minimum of $ERR_{edge}$ is attained for a high number of iterations. On the other hand, higher values of $\gamma_r$ require less iterations to attain the minimum but lead to an optimum with a higher error. This choice of values also allows to show that the global error $ERR_{glob}$ decreases with the number of iterations and that all the curves converge to the same value. Regarding the $ENL$, increasing both $\gamma_r$ and $N_{it}$ leads to higher values.
It may be observed that both Riemannian distances $d_{ai}$ and $d_{le}$ behave in a very similar way. To obtain comparable curves for all the distances, we show the results of the BLF with $d_{kl}$ with a different set of parameters. This difference can be explained in an intuitive way by considering the simpler case of two real positive scalars: $d_{kl}$ varies as the sum of the ratio and its inverse whereas $d_{le}$ and $d_{ai}$ behave as the logarithm of the ratio. Therefore, the two kinds of distances have different dynamics. 
The curves representing the ENL show that the BLF with Riemannian distances achieve a higher amount of smoothing in homogeneous areas.

According to this experiment, we select a set of parameters that correspond to a high ENL and good edge preservation while allowing a reasonable computational cost (computation time is mostly linked to the number of iterations.)
For $d_{ai}$ and $d_{le}$ we choose $\gamma_r = 1.33$ and $N_{it} = 4$ instead of $\gamma_r = 1.22$ and $N_{it} = 9$ because the gain in $ERR_{edge}$ and $ERR_{glob}$ is small and running the filter with 9 iterations would dramatically increase the compuational cost. Nevertheless, it is shown in the following that those slightly sub-optimal values are sufficient so that the BLF outperforms the other compared methods. 
Therefore, the selected parameter set for $d_{ai}$ and $d_{le}$ is $\gamma_r = 1.33$ and $N_{it} = 4$ while for $d_{kl}$ we choose $\gamma_r = 3.11$ and $N_{it} = 4$. 

Fig. \ref{fig:paraminfl} allows a visual comparison of the filtered image for the selected parameters and other parameter values. These images show that for the selected set of parameters, all edges are well-preserved in the image with a high level of smoothing on homogeneous areas. On the contrary, low values of $\gamma_r$ and $N_{it}$ result in a poor speckle reduction while too high values lead to blurry edges in low contrast areas. This visually confirms the measurements given in Fig. \ref{fig:curves}

\subsection{Validation on rank-1 targets.}
\label{sscn:exprank1}
It is important to verify that the method allows to avoid the filtering of purely deterministic targets. As explained in the previous section, ideal deterministic targets are unlikely to be present in experimental data. However, the presence of matrices that are close to rank 1 may result in numerical instability due to the use of matrix inversion and logarithm in the computation of the distances. Therefore, we have simulated a small image containing two ideal deterministic targets that correspond to rank one covariance matrices. A dot representing a purely trihedral target (the ideal representation of a corner reflector) is simulated with a diagonal matrix $\mathbf{T}_{tri} \propto diag{(1,0,0)}$ and a line modelling an ideal dihedral scatterer with matrix $\mathbf{T}_{di} \propto diag{(0,1,0)}$. The background is a homogeneous area with speckle simulated by the previously described method (see section \ref{sscn:SimVal}). We filtered this image with the BLF using the parameters selected in the previous section. Fig. \ref{fig:rank1} shows the results obtained for the BLF with $d_{ai}$ that involves both matrix inversion and logarithm. Similar results were obtained with the other versions of the filter. It can be noted that, thanks to the modification described in section \ref{sub:impl} to handle deterministic targets, the background is well smoothed and both pure targets are preserved. This results in correct $H/\alpha$ parameters for both background and rank 1 targets.

\subsection{Comparison with other filters}
\label{sub:compare}
We first have visually compared the three versions of the BLF for the previously selected parameters with the boxcar filter with a $7\times 7$ window, the refined Lee filter with $7$ pixel wide oriented windows and the IDAN filter for a neighborhood of $50$ pixels. We used implementations available in the RAT \cite{Rat} free software for the last two filters.

Fig. \ref{fig:simfilt} shows a visual comparison of the results obtained with the different methods on the synthetic image. For a better visualization areas of interest have been extracted from the original image.
While the boxcar filter blurs edges, degrading polarimetric information around boundaries, the BLF leads to sharp edges and smoothed estimates in homogeneous areas.
Again, we notice that the Riemannian distance based filters perform better than the Kullback-Leibler divergence based filter, that produces noisier estimates. This could be explained by the use of matrix logarithms in the Riemannian distances that make these quantities less sensitive to speckle. Although the refined Lee and IDAN filters allow a good preservation of edges, they also result in a noisier image. In the case of the Lee filter this is due to the use of a finite number of oriented windows. In the case of IDAN, this is due to the hard decision of including or not a pixel in the average. This effect is not present in the BLF filtered images thanks to the use of continuously varying weights.

Table \ref{fig:comperr} displays the values of the previously defined quality measures for all the filtering methods. All versions of the BLF lead to lower error than the other filters in terms of global reconstruction and edge preservation. Moreover, the BLF outperforms the other methods in terms of ENL. Furthermore, it may be observed that the BLF with Riemannian distances leads to the best estimates.

To evaluate the effects of the BLF on polarimetric information, it is useful to apply a polarimetric decomposition to the filtered data and to compare it to the ideal values. In this work, we choose to consider the entropy-alpha ($H/\alpha$) decomposition \cite{Cloude1996}. Here, $\alpha$ denotes the mean alpha angle.
For PolSAR data, it is also important that the filter does not lead to a systematic bias in the values of the covariance elements. On table \ref{fig:compcoh} we display the mean values computed over 4 manually selected homogeneous areas (shown in Fig. \ref{fig:simul}). We also show the mean of $\alpha$ angle and entropy $H$. The estimated values are close to the real ones for all coherency elements, including off-diagonal, as well as for the mean alpha angle and polarimetric entropy. It may be observed that the various versions of the BLF compare well with the other techniques.
This study also shows that both Riemannian versions of the BLF outperform the Kullback-Leibler based version.


\subsection{Experimental data}

\begin{figure*}[t]
\centering
\subfigure[4-look experimental image (crop).]{
\includegraphics[scale=0.15, angle=270]{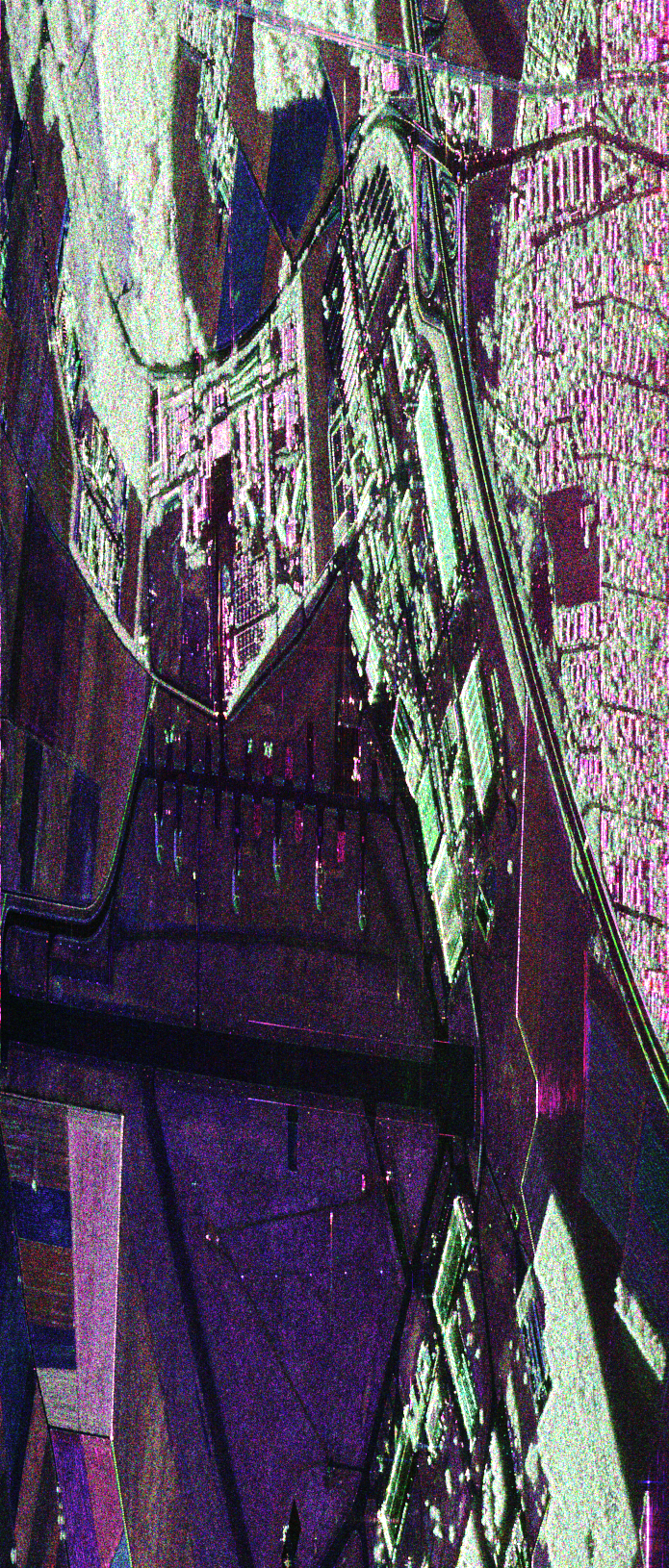}
}
\subfigure[BLF $d_{ai}$, $\gamma_r = 1.3$, $N_{it} = 4$]{
\includegraphics[scale=0.15, angle=270]{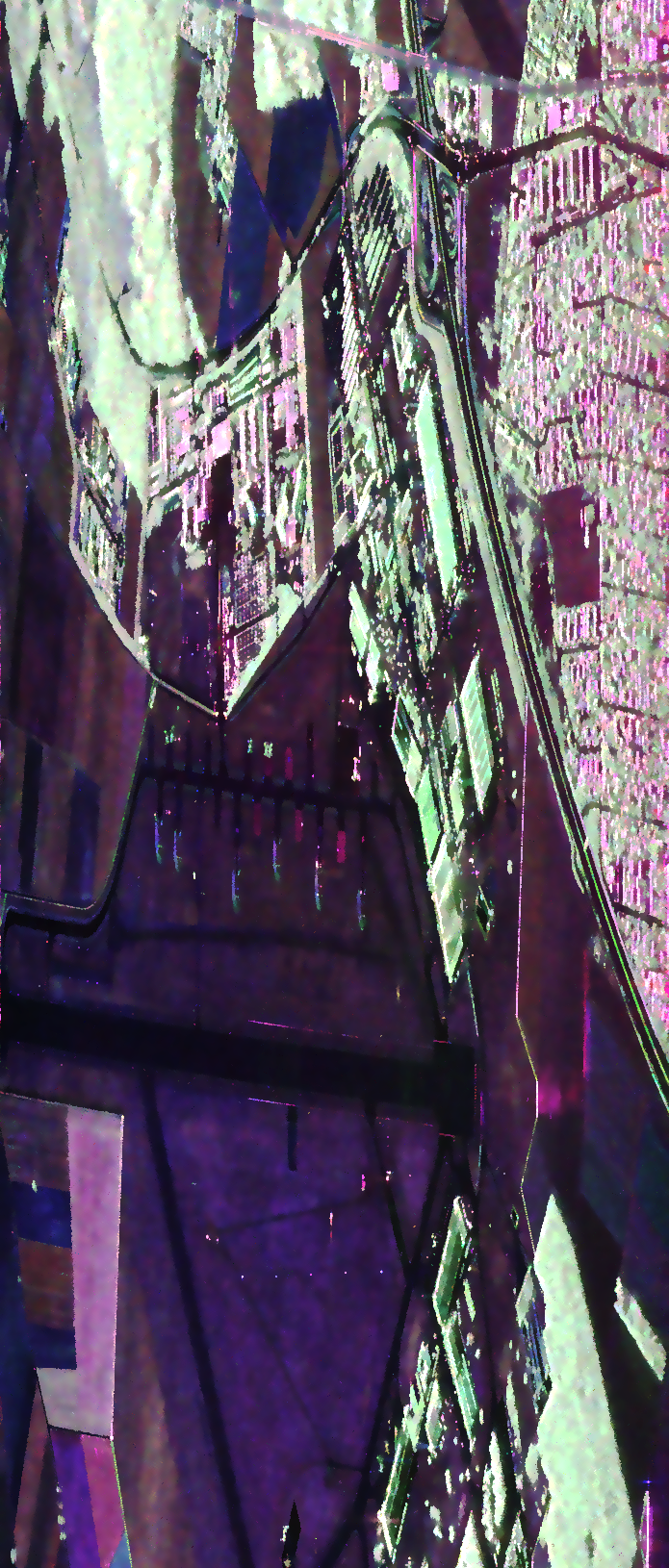}
}
\caption{Results of the BLF  with $d_{ai}$ applied to the Oberpfaffenhofen (DLR) dataset. BLF reduces speckle whithout blurring edges.}
\label{fig:ober}
\end{figure*}

\begin{figure*}[t]
\centering
\subfigure[Original (crop)]{
\includegraphics[scale=0.5]{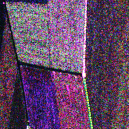}
\label{sfg:oc1a}
}
\subfigure[Boxcar $7\times7$]{
\includegraphics[scale=0.5]{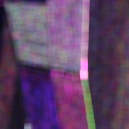}
\label{sfg:oc1b}
}
\subfigure[BLF $d_{ai}$]{
\includegraphics[scale=0.5]{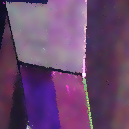}
\label{sfg:oc1c}
}
\subfigure[Refined Lee]{
\includegraphics[scale=0.5]{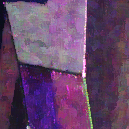}
\label{sfg:oc1d}
}
\subfigure[IDAN]{
\includegraphics[scale=0.5]{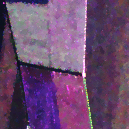}
\label{sfg:oc1e}
}\\
\subfigure[Boxcar $7\times7$]{
\includegraphics[scale=0.5]{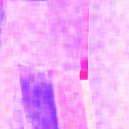}
\label{sfg:oc1f}
}
\subfigure[BLF $d_{ai}$]{
\includegraphics[scale=0.5]{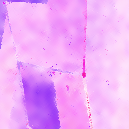}
\label{sfg:oc1g}
}
\subfigure[Refined Lee]{
\includegraphics[scale=0.5]{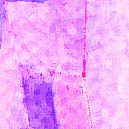}
\label{sfg:oc1h}
}
\subfigure[IDAN]{
\includegraphics[scale=0.5]{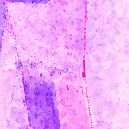}
\label{sfg:oc1i}
}
\subfigure[Color map]{
\includegraphics[scale=0.33]{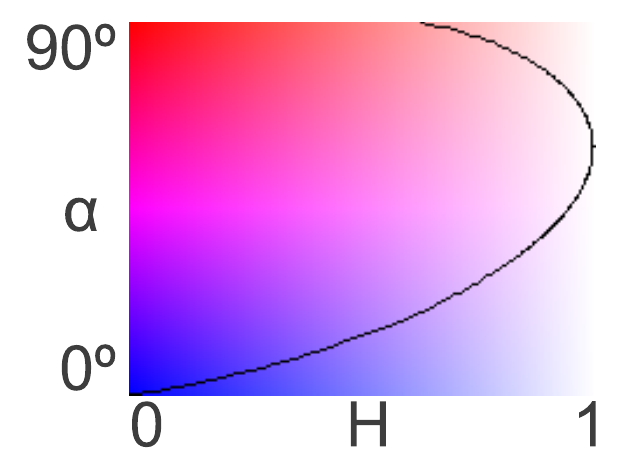}
\label{sfg:oc1j}
}

\caption{Results over a cropped area \subref{sfg:oc1a} from image of Fig. \ref{fig:ober}. The boxcar filter blurs the image \subref{sfg:oc1b} , the BLF leads to sharp edges and highly smoothed homogeneous areas \subref{sfg:oc1c}. Results with refined Lee \subref{sfg:oc1d} and IDAN \subref{sfg:oc1e} are displayed for comparison. The corresponding $H/\alpha$ parameters are displayed (\subref{sfg:oc1f} to \subref{sfg:oc1i}) with a Hue/Saturation  color map \subref{sfg:oc1j} .}
\label{fig:obercrop1}
\end{figure*}

\begin{figure}[t]
\centering
\subfigure[Original]{
\includegraphics[scale=1]{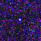}
\label{sfg:oc2a}
}
\subfigure[BLF $d_{ai}$]{
\includegraphics[scale=1]{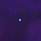}
\label{sfg:oc2b}
}
\subfigure[Ref. Lee]{
\includegraphics[scale=1]{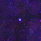}
\label{sfg:oc2c}
}
\subfigure[IDAN]{
\includegraphics[scale=1]{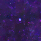}
\label{sfg:oc2d}
}\\
\subfigure[BLF $d_{ai}$]{
\includegraphics[scale=1]{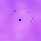}
\label{sfg:oc2e}
}
\subfigure[Ref. Lee]{
\includegraphics[scale=1]{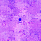}
\label{sfg:oc2f}
}
\subfigure[IDAN]{
\includegraphics[scale=1]{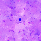}
\label{sfg:oc2g}
}
\subfigure[Color map]{
\includegraphics[scale=0.33, trim = 0.2cm 0.2cm 0 0, clip=true]{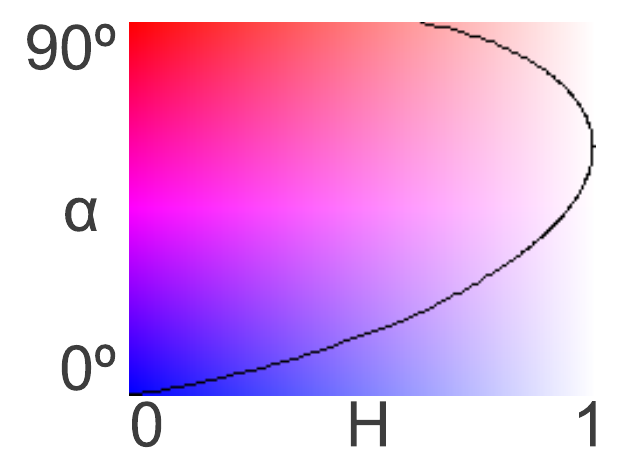}
\label{sfg:oc2h}
}
\caption{Results over a cropped area from image of Fig. \ref{fig:ober} with a corner reflector and a homogeneous background \subref{sfg:oc2a}. The BLF  \subref{sfg:oc2b} allows a good restoration the point target with a smooth background whereas the other filters \subref{sfg:oc2c}, \subref{sfg:oc2d} result in a noisier result. The corresponding $H/\alpha$ parameters are shown  \subref{sfg:oc2e}, \subref{sfg:oc2f}, \subref{sfg:oc2g} with a Hue-Saturation color map \subref{sfg:oc2h}.}
\label{fig:obercrop2}
\end{figure}

\begin{figure}[t]
\centering
\subfigure[Boxcar $7{\times}7$]{
\includegraphics[scale=0.4, trim=2.1cm 0.3cm 2.9cm 0.8cm, clip=true]{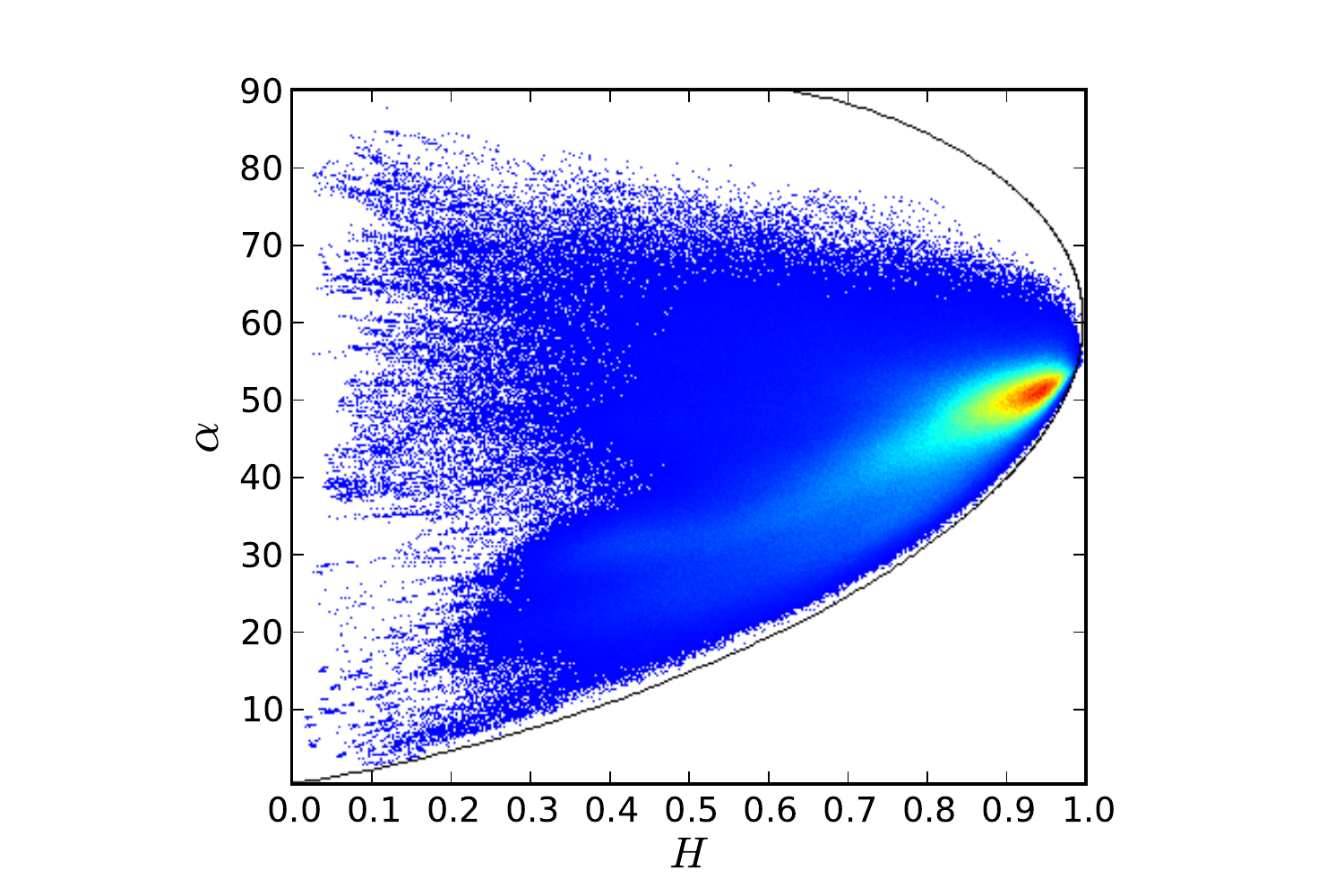}
\label{sfg:haa}
}
\subfigure[BLF $d_{ai}$]{
\includegraphics[scale=0.4, trim=2.1cm 0.3cm 2.9cm 0.8cm, clip=true]{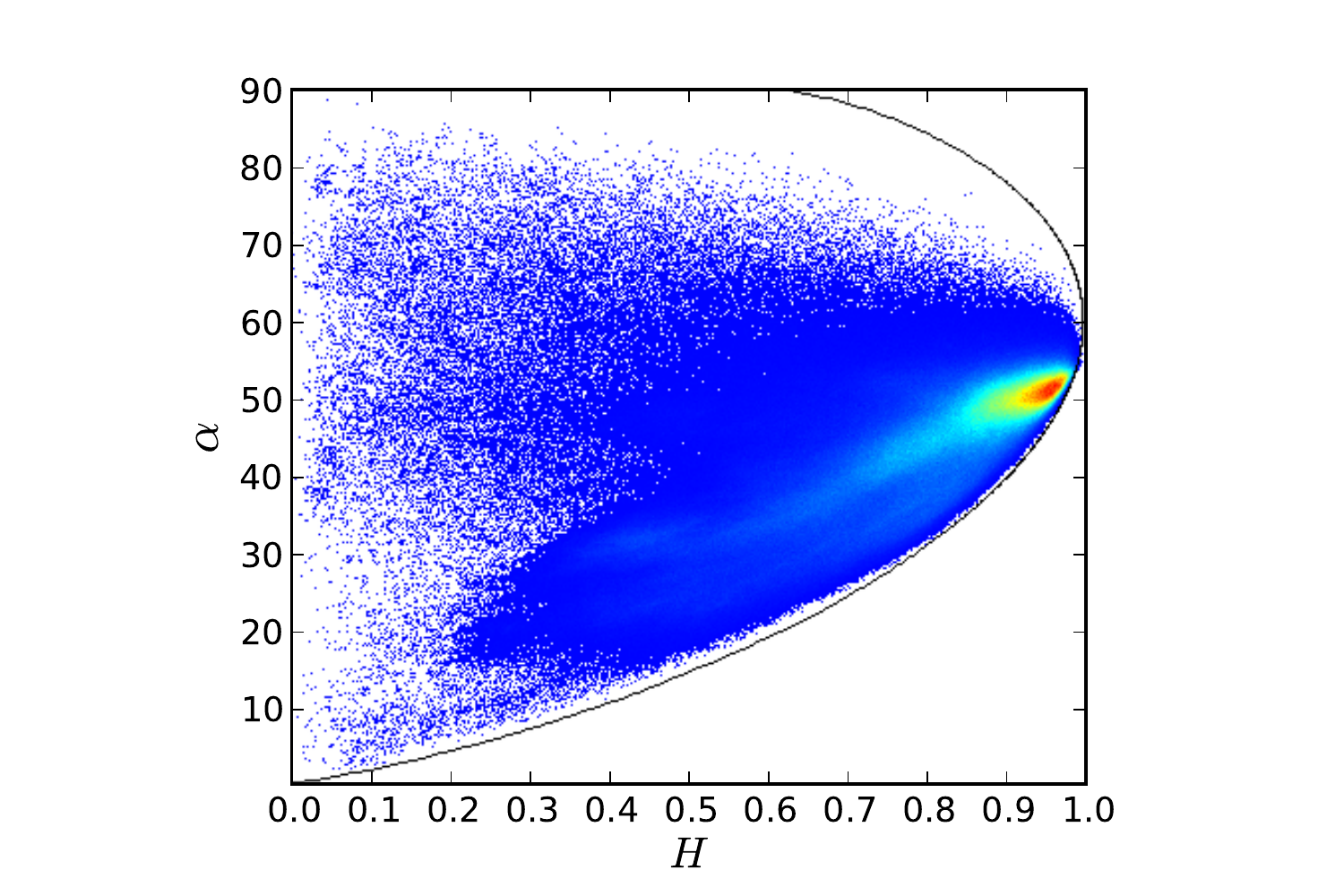}
\label{sfg:hab}
}\\
\subfigure[Refined Lee]{
\includegraphics[scale=0.4, trim=2.1cm 0.3cm 2.9cm 0.8cm, clip=true]{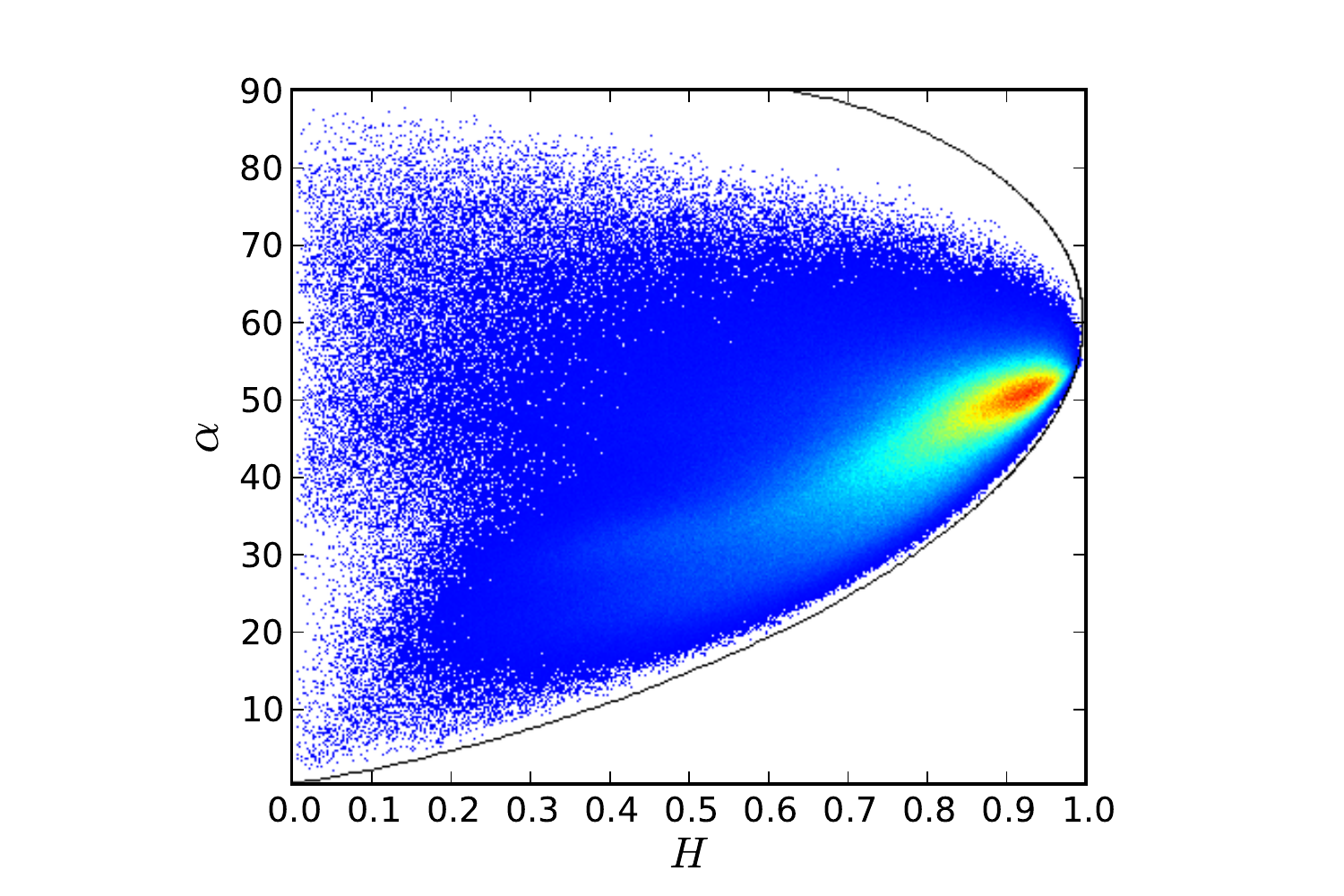}
\label{sfg:hac}
}
\subfigure[IDAN]{
\includegraphics[scale=0.4, trim=2.1cm 0.3cm 2.9cm 0.8cm, clip=true]{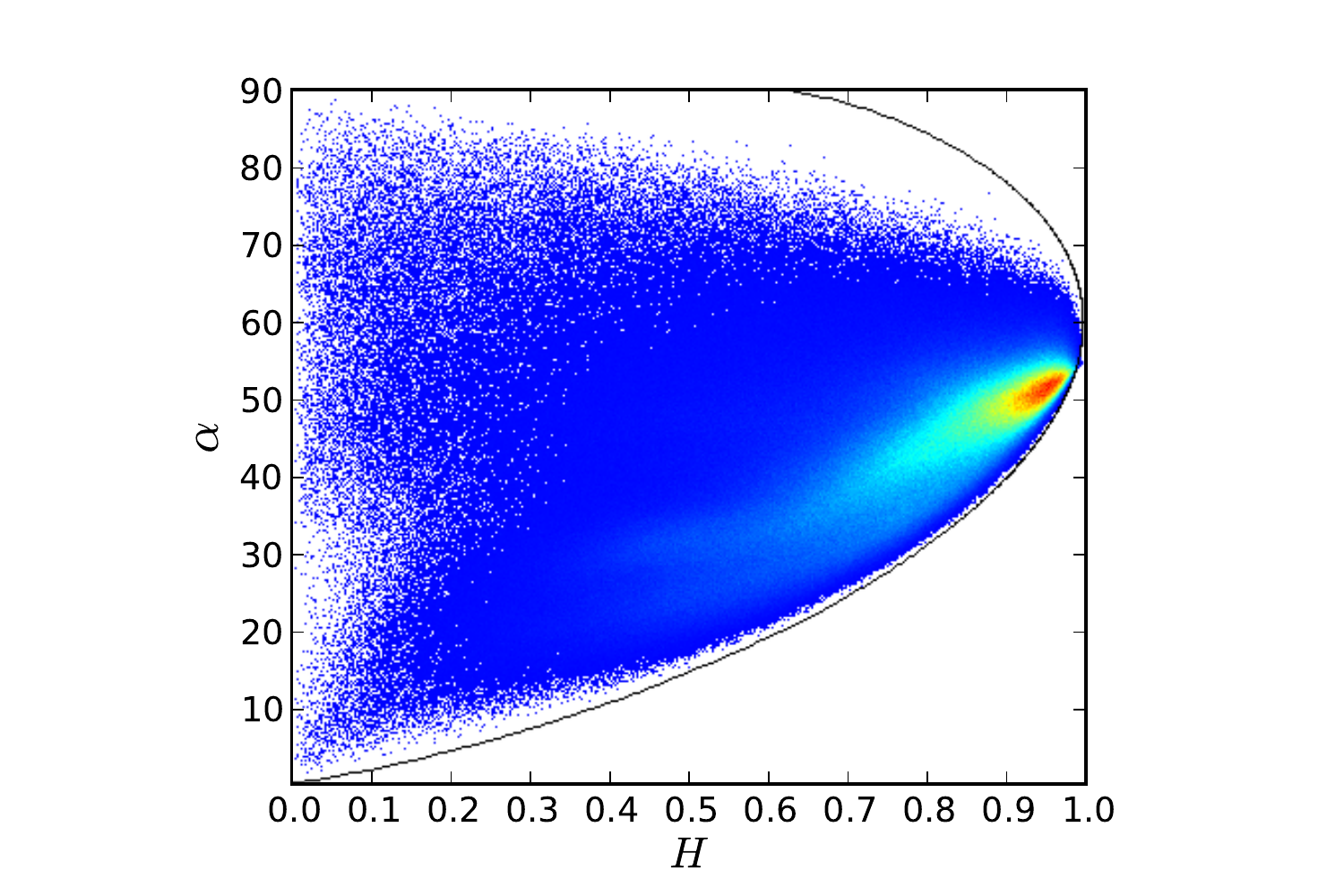}
\label{sfg:had}
}\\
\subfigure{
  \includegraphics[scale=0.4]{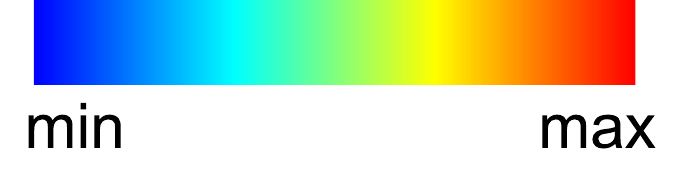}
\label{sfg:hae}
}
\caption{Comparison of point densities in the $H/\alpha$ plane for $7 {\times} 7$ boxcar \subref{sfg:haa}, BLF with $d_{ai}$ \subref{sfg:hab}, refined Lee \subref{sfg:hac} and IDAN \subref{sfg:had}.}
\label{fig:halphaplane}
\end{figure}

\begin{figure}[t]
\centering
\subfigure[Original FSAR image (S-Band)]{
\includegraphics[scale=0.2]{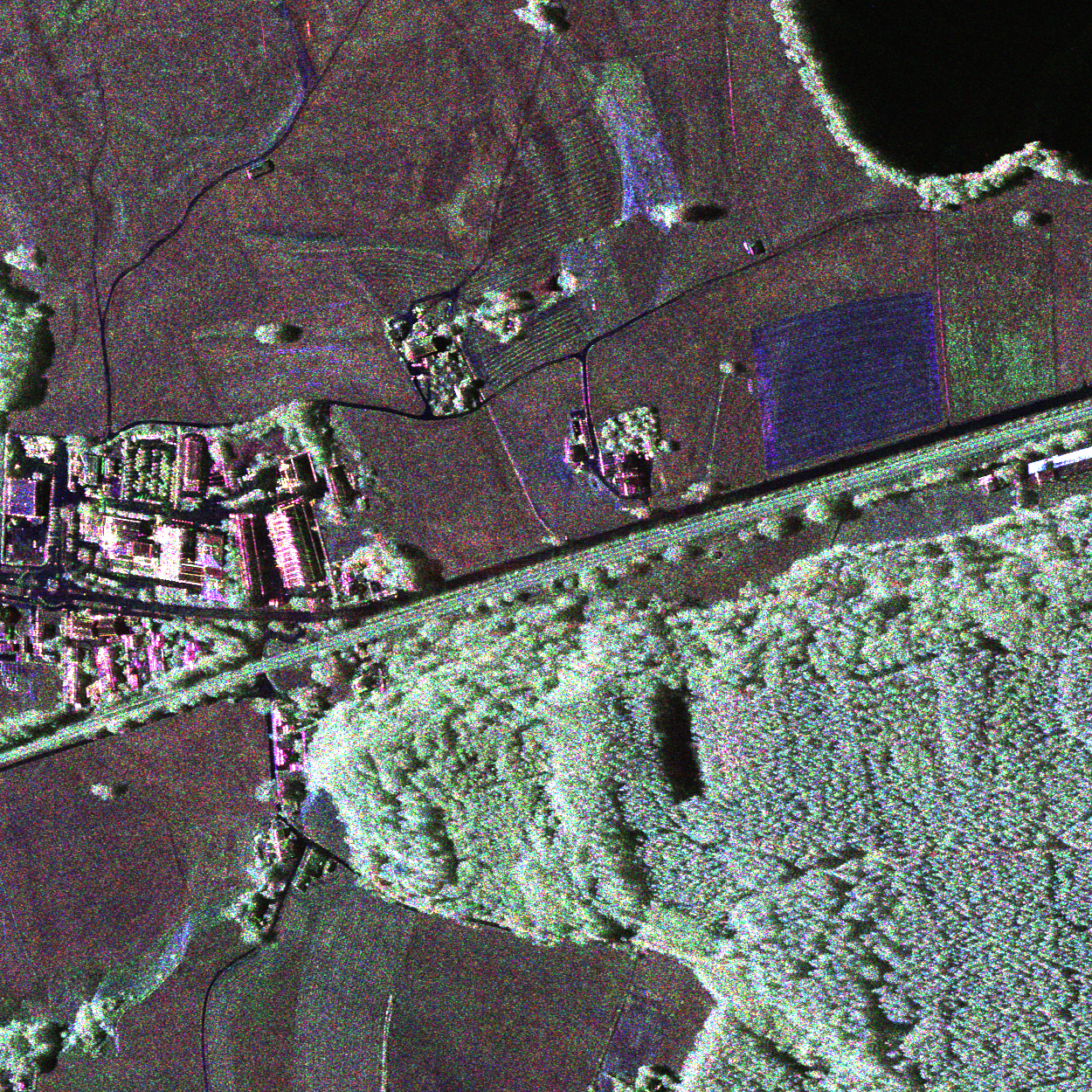}
}
\subfigure[Filtered with BLF $d_{ai}$, $\gamma_r=1.33$, $N_{it} = 4$]{
\includegraphics[scale=0.2]{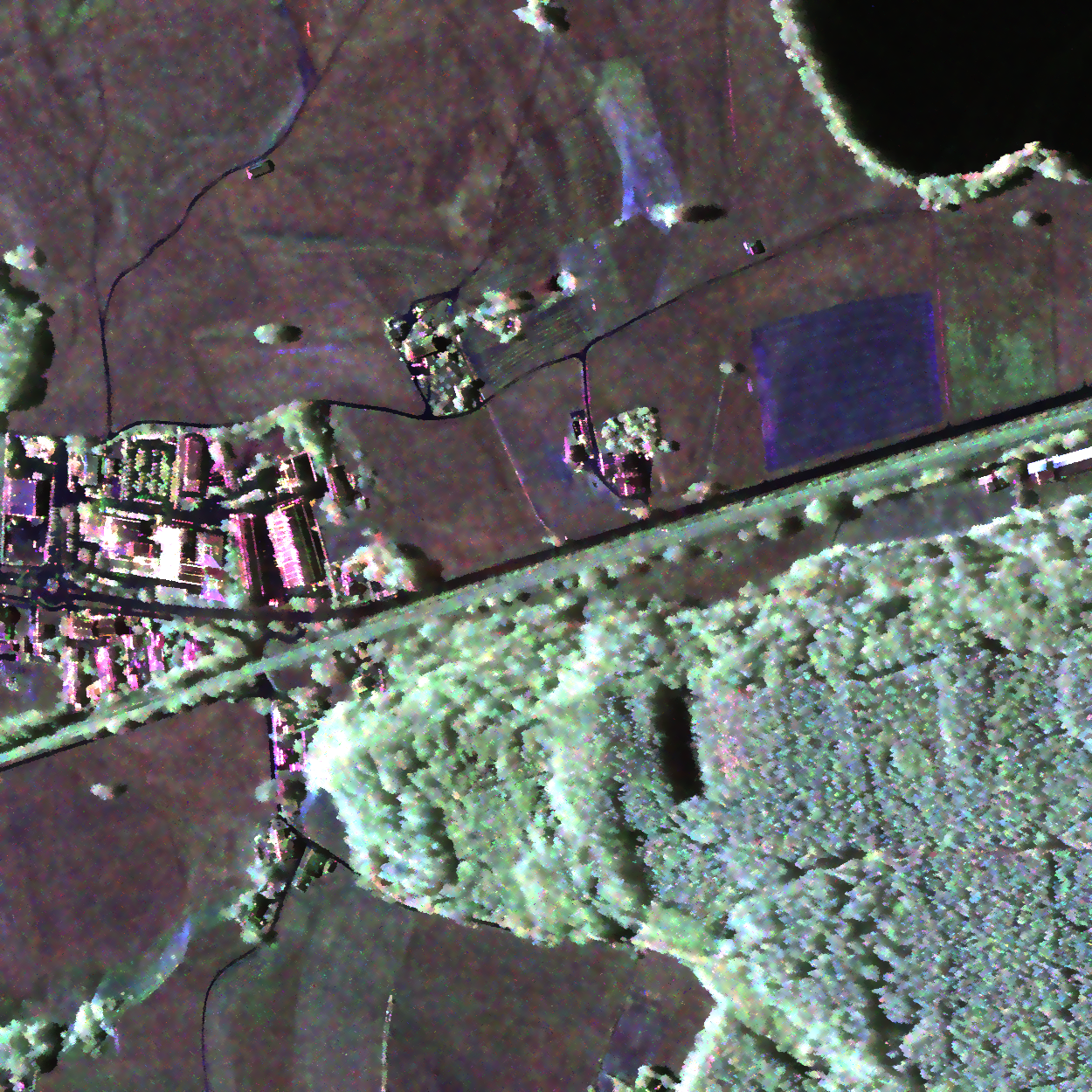}
}
\caption{Filtering results for the S-Band FSAR demo image (DLR). Speckle is reduced and structures such as buildings and roads are well preserved. Spatial texture is also preserved in the forested area.}
\label{fig:fsar}
\end{figure}

Fig. \ref{fig:ober} shows the result of the BLF with $d_{ai}$, applied to the experimental Oberpfaffenhofen dataset introduced at the beginning of the section.
For a better visualization, only a cropped area of the original image is shown.
This experiment was also done with $d_{le}$ that showed similar performance. As a preprocessing, a pre-summing of 2 pixels in both azimuth and range has been applied.
The filtering parameters are the ones retained in the previous sections.
It can be noted that the filter is able to preserve complex spatial structures in urban and forested areas and strongly reduces speckle on homogeneous areas. Fig. \ref{fig:obercrop1} shows a zoom on a sub-image with several homogeneous areas and compares the BLF with the boxcar, refined Lee and IDAN filters. The BLF results in sharp edges and very smooth homogeneous areas. As observed in simulations, the refined Lee and IDAN filters also preserve edges but they also lead to a noisier image that can be a problem for applications such as segmentation or edge detection. The drawback of the BLF is a slight over-smoothing of very low contrast edges. A representation of the estimated alpha entropy parameters is shown in the bottom row, using a Hue/Saturation color map. If the $H/\alpha$ parameters look roughly the same for all the filters, the BLF filtered image has a smoother aspect than the other ones.
Fig. \ref{fig:obercrop2} shows a zoom on a corner reflector. It may be observed that the BLF allows a good reconstruction of the reflector, and performs a better background smoothing than the other methods.

Fig. \ref{fig:halphaplane} shows representations of the point density in the  $H/\alpha$ plane for boxcar, BLF, Lee and IDAN. 
Both BLF and boxcar lead to more compact scatter plots than Lee and IDAN filter, suggesting a higher amount of averaging. However, the BLF point cloud contains more very low entropy points than the boxcar one, suggesting its ability to preserve those points whereas boxcar tends to mix them with other higher entropy points.

Finally, we show results obtained with the same set of parameters on the more recent FSAR S-Band demo set provided by DLR (Fig. \ref{fig:fsar}). The BLF achieves a high-quality reconstruction of the different types of structures in the image. In particular, it may be observed that buildings and roads are visually well preserved, as well as spatial texture in forested areas. 

\section{Conclusion}
\label{sec:Conclusion}
In this paper, we have developed a new speckle filter for PolSAR data based on an iterative version of the bilateral filter. To take advantage of the radiometric similarities between covariance matrices in a local neighbourhood, we have used weights based on suitable distances.

The Kullback-Leibler statistical distance has been compared to two Riemannian distances, that are defined on the manifold of Hermitian positive definite matrices.

We have applied this filter to simulated and experimental data. An in-depth evaluation has been performed over synthetic data. A study of the results obtained by all versions of the filter with different sets of parameters has been achieved. This study allowed a selection of suitable parameters for the comparison with other filters. The BLF showed good performance for spatial preservation of edges and deterministic targets. Moreover, thanks to an iterative scheme, the BLF outperforms other methods in terms of ENL. The preservation of the $H/\alpha$ polarimetric parameters is also achieved. The best performance has been obtained with the Riemannian distance based filters. Finally, our filter has been applied to experimental data and achieved a high-quality restoration of complex spatial structures.

In this work we have considered only fully-developed speckle that corresponds to a Gaussian distribution of target vectors. To apply such an approach to very high resolution data, it should also be studied in the case of non-Gaussian data. Effects of the filter on spatially correlated texture should also be considered in future work. Since, the use of Riemannian based similarities showed the best performance, it would also be interesting to investigate the relation between Riemannian means and the parameters of probability distributions such as Wishart or K-distribution. In order to avoid a manual tuning of the parameters, automatic selection should be addressed in future research. Finally, the performance of the filter in terms of sharp edge reconstruction and smoothing of homogeneous areas suggest to investigate its use in applications such as segmentation, edge detection and machine learning.

\section*{Acknowledgement}
The authors would like to thank the anonymous reviewers for their constructive comments that greatly helped improving the quality of the paper.

The authors also would like to thank DLR for providing the FSAR and ESAR datasets.
This work was supported by the German Science Foundation DFG, under project No. HE 2459/15-1

\bibliographystyle{IEEEtran}
\bibliography{BibJSTARS2012}
\begin{IEEEbiography}[{\includegraphics[width=1in]{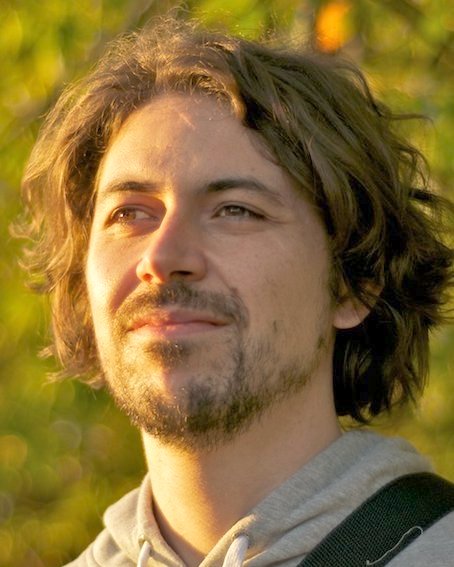}}]
{Olivier D'Hondt} received the M.S. degree in electrical engineering in 2002 and the Ph.D. degree in Signal Processing and Telecommunication in 2006, both from the University of Rennes 1, France. 
From March 2006 to October 2007, he was a Postdoctoral Fellow and stayed as an invited member with the Remote Sensing Laboratory (RSLab), Universitat Polit\`{e}cnica de Catalunya (UPC) in Barcelona, Spain. He also was with the Geophysical Imagery team from the Geosciences Rennes Laboratory, University of Rennes 1, France doing research on nonstationary texture analysis in SAR and sidescan SONAR images. 
From  November 2007 to January 2011, he was a research scientist at the Barcelona Media Research Center in Barcelona, Spain, working on video analysis for visual post-production.
Since September 2011 he is with the Computer Vision and Remote Sensing Group at  Technische Universit\"{a}t Berlin (TUB), Berlin, Germany. His current research activity deals with combining extraction of physical parameters and object detection to improve multi-channel SAR image analysis.
He received a Student Paper Award at the European Conference in Synthetic Aperture Radar (EUSAR) 2004, Ulm, Germany for the paper entitled Non Stationary Texture Analysis from Polarimetric SAR Data.
\end{IEEEbiography}

\begin{IEEEbiography}[{\includegraphics[width=1in]{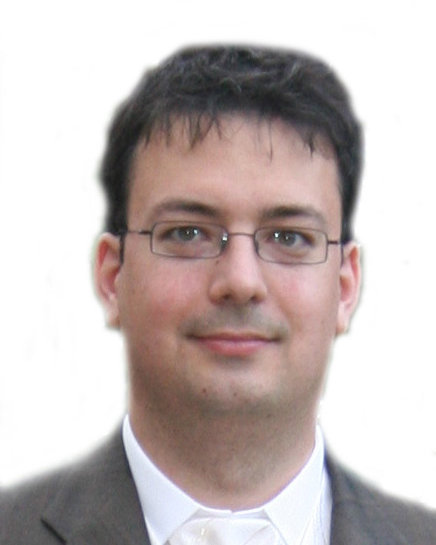}}]
{St\'{e}phane Guillaso} was born in Marseille, France. After studying applied physic at the "Universit\'{e} de la M\'{e}diterran\'{e}e", Marseille, France from 1994 to 1999, he received the M.Sc. and the Ph.D. degrees in signal processing and telecommunication from the University of Rennes 1, France, in 2000 and 2003, respectively.
In 2000, he was with the Microwaves and Radar Institute, DLR, Oberpfaffenhofen, Germany to work on the improvement of range resolution of airborne SAR images. From 2000 to 2003, he was with the SAR Polarimetry, Holography, Interferometry and Radargrammetry (SAPHIR) Team, Institute of Electronics and Telecommunications of Rennes, University of Rennes 1, France, to work on the complementarity of polarimetry and interferometry. In 2004 he joined the Computer Vision and Remote Sensing Laboratories, Berlin University of Technology, Berlin, to work on polarimetric SAR tomography. From 2007 to 2009 he was with the Geology Laboratory of the Ecole National Sup\'{e}rieure, Paris, France, to work on the differential SAR interferometry. Sing 2009 he has been with the Computer Vision and Remote Sensing, Berlin University of Technology, Berlin, Germany, where he is currently Research Associate. He is currently working on different methodological aspects of processing and understanding multimodal, multitemporal SAR images.
Dr. Guillaso was the recipient of the 2007 IEEE GRS-S LETTER PRIZE PAPER AWARD. 
\end{IEEEbiography}

\begin{IEEEbiography}[{\includegraphics[width=1in]{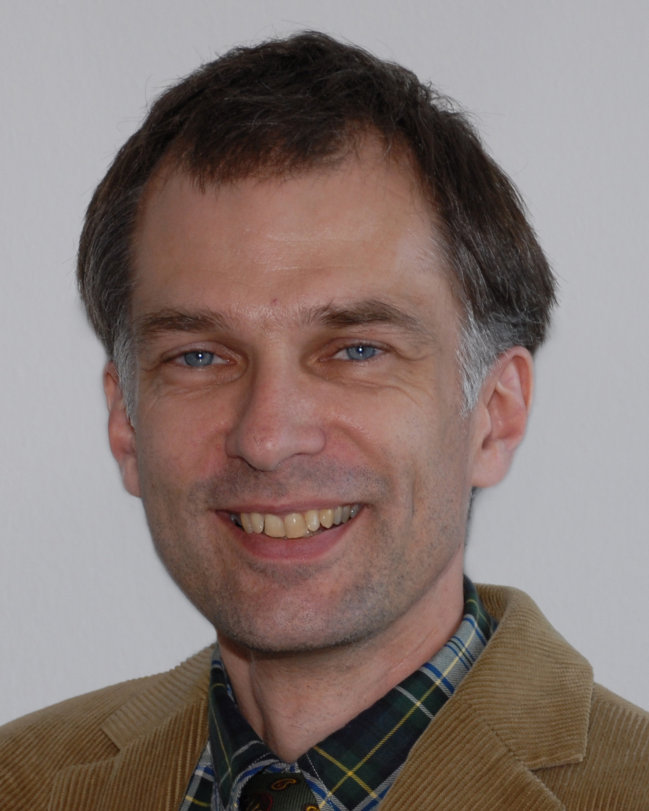}}]
{Olaf Hellwich} (M’98–SM’06) was born in 1962. He received the B.S. degree in surveying engineering
from the University of New Brunswick, Fredericton, NB, and the Ph.D. degree from the Technische Universit\"{a}t M\"{u}nchen, M\"{u}nchen, Germany, in 1997.
He headed the Remote Sensing Group, Department of Photogrammetry and Remote Sensing, Technische Universit\"{a}t M\"{u}nchen. Since 2001, he has been a Professor with the Technische Universit\"{a}t Berlin (TUB), Berlin, Germany, initially for photogrammetry and cartography and since 2004 for computer vision and remote sensing. From 2006 to 2009 he was Dean of the Faculty of Electrical Engineering and Computer Science, TUB. His research interests are in 3-D object reconstruction, e.g., from video sequences, object recognition, e.g., real-time head pose estimation, and synthetic aperture radar remote sensing, e.g., for surface motion estimation.
Dr. Hellwich was the recipient of the Hansa Luftbild Prize of the German Society for Photogrammetry and Remote Sensing in 2000.
\end{IEEEbiography}

\end{document}